\documentclass[journal]{IEEEtran}
\usepackage{amsmath,amsfonts}
\usepackage{algorithm}
\usepackage{algpseudocode}
\usepackage{color}
\usepackage{soul}
\usepackage{cite}
\usepackage{graphicx}
\usepackage{amsthm}
\usepackage{multirow}
\def\BibTeX{{\rm B\kern-.05em{\sc i\kern-.025em b}\kern-.08em
    T\kern-.1667em\lower.7ex\hbox{E}\kern-.125emX}}
\usepackage{subfigure}
\begin{document}
\title{Decentralized iLQR for Cooperative Trajectory Planning of Connected Autonomous Vehicles via Dual Consensus ADMM}
\author{Zhenmin Huang, Shaojie Shen, and Jun Ma
\thanks{Zhenmin Huang, Shaojie Shen, and Jun Ma are with the Department of Electronic and Computer Engineering, The Hong Kong University of Science and Technology, Hong Kong SAR, China (email: zhuangdf@connect.ust.hk; eeshaojie@ust.hk; jun.ma@ust.hk)}
\thanks{This work has been submitted to the IEEE for possible publication. Copyright may be transferred without notice, after which this version may no longer be accessible.}
}

\markboth{}{}
\maketitle

\begin{abstract}
Developments in cooperative trajectory planning of connected autonomous vehicles (CAVs) have gathered considerable momentum and research attention. Generally, such problems present strong non-linearity and non-convexity, rendering great difficulties in finding the optimal solution. Existing methods typically suffer from low computational efficiency, and this hinders the appropriate applications in large-scale scenarios involving an increasing number of vehicles. To tackle this problem, we propose a novel decentralized iterative linear quadratic regulator (iLQR) algorithm by leveraging the dual consensus alternating direction method of multipliers (ADMM). First, the original non-convex optimization problem is reformulated into a series of convex optimization problems through iterative neighbourhood approximation. Then, the dual of each convex optimization problem is shown to have a consensus structure, which facilitates the use of consensus ADMM to solve for the dual solution in a fully decentralized and parallel architecture. Finally, the primal solution corresponding to the trajectory of each vehicle is recovered by solving a linear quadratic regulator (LQR) problem iteratively, and a novel trajectory update strategy is proposed to ensure the dynamic feasibility of vehicles. With the proposed development, the computation burden is significantly alleviated such that real-time performance is attainable. Two traffic scenarios are presented to validate the proposed algorithm, and thorough comparisons between our proposed method and baseline methods (including centralized iLQR, IPOPT, and SQP) are conducted to demonstrate the scalability of the proposed approach. 
\end{abstract}


\theoremstyle{definition}
\newtheorem{definition}{Definition}
\theoremstyle{definition}
\newtheorem{remark}{Remark}
\theoremstyle{definition}
\newtheorem{assumption}{Assumption}
\theoremstyle{definition}
\newtheorem{lemma}{Lemma}
%\graphicspath{ {pictures/} }

\section{Introduction}

The continuous increase in the number of on-road vehicles has imposed a heavy burden on the traffic system, resulting in severe congestion and safety issues. Meanwhile, the generally non-cooperative nature of human drivers leads to intense competition for limited traffic resources and even worsens the situation. The existing issues, together with the rapid development of autonomous driving technologies and information technologies, prompt the emergence of connected autonomous vehicles (CAVs), which provide a cooperative driving solution that can potentially alleviate the problem~\cite{sun2021survey}. CAVs are generally equipped with one or more communication devices that enable the information exchange with other on-road vehicles (vehicle-to-vehicle), roadside infrastructure (vehicle-to-infrastructure), and cloud device (vehicle-to-cloud)~\cite{chehri2020communication}. Such communication conveys the driving intention between CAVs and forms the physical basis of cooperative driving. However, from the perspective of algorithm development, the problem of cooperative trajectory planning between CAVs still remains largely unsolved. The main target of cooperative trajectory planning is to make joint decisions and generate cooperative trajectories for all involved CAVs that satisfy vehicle dynamics, traffic rules, safety conditions, and other pertinent constraints while maximizing the overall efficiency~\cite{xu2018cooperative,hang2020integrated,hang2021cooperative}. Although various methods are proposed, the complexity of traffic scenarios and the inherent coupling nature of cooperative trajectory planning render this kind of problem hard to solve. Moreover, with the increasing number of participant CAVs, the scale of the cooperative trajectory planning problem also expands, resulting in deteriorated time performance. These issues pose great difficulties in finding the optimal solution in real-time, especially under limited computational resources and for large-scale scenarios, making cooperative trajectory planning a long-standing challenge in the scope of autonomous driving.

Currently, two categories of methods are intensively investigated to tackle the cooperative trajectory planning problem. The learning-based methods provide a feasible way to leverage real-world driving data and capture complicated driving behaviours to handle difficult traffic scenarios~\cite{guan2020centralized,chu2019multi,li2022decision}. However, the requirement for an excessive amount of real-world data and the lack of interpretability of neural networks hinder their wide applications. On the other hand, the optimization-based methods are more mature and well-developed with predictable results and good interpretability. Such methods formulate the cooperative trajectory planning problem in the form of constrained optimization, which incorporates prescribed constraints concerning safety conditions and control limits, and try to minimize pertinent objective functions such as control costs and deviation from reference track, etc. Within the family of optimization-based methods, both centralized approaches and decentralized approaches are well researched. The centralized approaches generally adopt a central coordinator that perceives complete information and makes decisions for all CAVs~\cite{wu2014cooperative,de2013autonomous}. A hierarchical centralized coordination scheme is proposed in~\cite{pan2022convex}, which is performed by a central traffic coordinator that handles the traffics at the intersection. Although centralized methods serve as a natural solution to the cooperative trajectory planning problem, one of the shortcomings is the low computational efficiency caused by the increasing number of CAVs and the limited computing power of the central coordinator. Such poor scalability with respect to the number of vehicles prevents their application to scenarios of large scale. On the contrary, decentralized methods require every CAV to make its own decision based on local information~\cite{campos2014cooperative,tedesco2010distributed,kim2014mpc}. Therefore, the computation load is naturally distributed among all vehicles, which results in better scalability. Nevertheless, incomplete information perceived by each CAV essentially complicates the optimization problem; and apparently, more efforts on algorithm development are required.

In general, the optimization problems formulated by optimization-based methods involve strong non-linearity. Therefore, nonlinear programming is typically required; and in this sense, the efficiency of which relies heavily on the solver. Sequential quadratic programming (SQP) and interior point optimizer (IPOPT) are widely used nonlinear programming solvers for solving such optimization problems, but they suffer from low efficiency in most scenarios. Compared with SQP and IPOPT, differential dynamic programming (DDP) presents desirable features such as low memory consumption and significant improvement in computational efficiency, and therefore triggers wide applications to various kinds of optimal control problems~\cite{pan2020safe,kong2022hybrid,ma2022local}. Meanwhile, the iterative linear quadratic regulator (iLQR) provides a simplified version of DDP by eliminating the second-order terms of system dynamics to further enhance the efficiency. The main obstacle that prevents the usage of the original DDP/iLQR algorithm in trajectory planning for vehicles is that it cannot handle constraints other than system dynamics. To solve this problem, control-limited DDP~\cite{tassa2014control} incorporates the control limits into the DDP framework, while CiLQR~\cite{chen2019autonomous} combines iLQR with log barrier function to further incorporate general forms of constraints. These works enable the application of DDP/iLQR to trajectory planning. The recent application of iLQR to cooperative trajectory planning is inspired by~\cite{kavuncu2021potential}, which formulates the interactive trajectory planning problem as a potential game and obtains the corresponding Nash equilibrium by solving a single optimal control problem with the iLQR algorithm. Although being one of the most efficient and scalable methods in the scope of optimal control and trajectory planning, the direct application of DDP/iLQR onto high-dimensional dense systems in a centralized fashion still results in suboptimal performance. The development of a distributed version of DDP/iLQR is desirable to further enhance the computational efficiency and scalability.

With its distributed nature, the alternating direction method of multipliers (ADMM)~\cite{boyd2011distributed} is widely deployed for solving optimization problems in various domains~\cite{raja2020sp,ma2022symmetric,cheng2020semi,zhang2018admm}. The basic idea of ADMM is to decompose the original problem into smaller, manageable ones such that they can be solved in a parallel and distributed manner. In~\cite{ma2022alternating}, a trajectory planning algorithm based on ADMM is presented to separate the vehicle dynamic constraints from other constraints such that they can be dealt with respectively; and superior performance in terms of computational efficiency over SQP and IPOPT is demonstrated. Meanwhile, consensus ADMM is proposed~\cite{mateos2010distributed} for solving the consensus optimization problem over a connected undirected graph. Due to the broad existence of systems with a communication network structure, it has been deployed in various domains such as resource allocation~\cite{doostmohammadian2022distributed} and model predictive control~\cite{aboudonia2022online}. This is followed by dual consensus ADMM~\cite{banjac2019decentralized,grontas2022distributed}, which extends the consensus ADMM to an optimization problem whose dual problem is a consensus optimization problem. The recent applications of ADMM to cooperative trajectory planning are exemplified by~\cite{cheng2021admm,zhang2021semi,zhang2021parallel,saravanos2022distributed}. Particularly in~\cite{cheng2021admm,zhang2021semi,zhang2021parallel}, ADMM is applied to separate the independent constraints from the coupling constraints of the multi-agent system, such that the former ones can be handled in a parallel manner, yielding a partially decentralized solution. Meanwhile, a fully decentralized optimization framework is proposed in~\cite{saravanos2022distributed} to solve multi-robot cooperative trajectory planning problems over a large scale. However, its performance is still far away from being real-time, even with a relatively small number of agents. 

Inheriting both the efficiency of iLQR towards non-linearity and the distributed nature of dual consensus ADMM, this paper presents a novel fully decentralized and parallelizable optimization framework for cooperative trajectory planning of CAVs, based on a convex reformulation methodology to tackle the inherent non-convexity residing in the corresponding optimization problem. Through the optimization framework, coupling constraints on safe distances between CAVs are essentially decoupled and handled by all CAVs collaboratively. Nonlinear vehicle dynamics and control limits are ensured to be satisfied by a novel dynamically feasible trajectory update strategy. The main contributions of this paper are listed as follows:

\begin{itemize}
\item[$\bullet$]
A convex reformulation methodology is proposed to address the non-convexity in the original constrained optimization problem for cooperative trajectory planning of CAVs. 
% Thus, the convergence of the proposed optimization framework is ensured.
\end{itemize}

\begin{itemize}
\item[$\bullet$]
A novel fully decentralized and parallelizable optimization framework is introduced by inheriting the merits of iLQR and dual consensus ADMM to split the original optimization problem into sub-problems, such that the computation load is distributed evenly among all CAVs and the computational efficiency is enhanced significantly.
\end{itemize}

\begin{itemize}
\item[$\bullet$]
An innovative dynamically feasible trajectory update strategy based on iLQR is introduced to guarantee the satisfaction of system dynamics and prescribed control limits of the CAVs.
\end{itemize}

\begin{itemize}
\item[$\bullet$]
A fully parallel implementation of the proposed method based on multi-processing is provided to reach real-time performance. 
% Different driving scenarios including T-junction and intersection. 
Since the proposed development is a unified framework for trajectory planning of multi-agent systems, the same implementation can be readily generalized to a wide range of applications.
\end{itemize}

The remainder of this paper is organized as follows. Section II presents the formulation of the cooperative trajectory planning problem for connected autonomous vehicles. Section III presents the convex reformulation and the decentralized iLQR algorithm based on dual consensus ADMM for solving the problem. In Section IV, two traffic scenarios are provided to demonstrate the effectiveness of the proposed algorithm, and thorough discussions are also presented. At last, Section V gives the conclusion of this work and several possible future works.

\textit{Notations:} $\mathbb{R}^{a\times b}$ denotes the space of real matrices containing $a$ rows and $b$ columns, and $\mathbb{R}^a$ to denote the space of $a$-dimensional real column vectors. Similarly, we use $\mathbb{Z}^{a\times b}$ to denote integer matrices containing $a$ rows and $b$ columns and $\mathbb{Z}^a$ to denote $a$-dimensional integer column vectors, respectively. $\textbf{0}^{a\times b}$ represents an all-zero matrix with $a$ rows and $b$ columns. $I^n$ denotes an $n$ by $n$ identity matrix. $A^{\top}$ and $x^{\top}$ represent the transpose of a matrix and a column vector, respectively. The operator $||\boldsymbol{\cdot}||$ denotes the Euclidean norm of a vector. We also use $\textup{blocdiag}\{A_1,A_2,...,A_n\}$ to denote the block diagonal matrix with block diagonal entries $A_1,A_2,...,A_n$. $(x_1,x_2,...,x_n)$ denotes the concatenated vector of the following sets of column vectors $\{x_1, x_2, ..., x_n\}$, namely $(x_1,x_2,...,x_n)=[x_1^{\top},x_2^{\top},...,x_n^{\top}]^{\top}$. The proximal operator is defined as $\textup{Prox}^\rho_f(x):=\underset{y}{\arg\min}\{f(y)+\frac{\rho}{2}||x-y||^2\}$.

\section{Problem Formulation}
For a traffic scenario involving $N$ CAVs, we use $\mathcal{N}=\{1,2,...,N\}$ to denote the set containing the index of each CAV. Considering the discrete-time setting, we use $\mathcal{T}=\{0,1,...,T-1\}$ to denote the time stamps. The system dynamics for each vehicle $i, i\in \mathcal{N}$, is given by the following discrete-time non-linear equations
\begin{equation}
x^i_{\tau+1}=f(x^i_\tau,u^i_\tau),
\end{equation}
where $x^i_\tau\in\mathbb{R}^n$ and $u^i_\tau\in\mathbb{R}^m$ are the state vector and control input vector of vehicle $i$ at time $\tau$ for $\tau\in\mathcal{T}$, and $x^i_0$ is given and fixed. 

We denote $x_\tau = (x^1_\tau,x^2_\tau,...,x^N_\tau)$ and $u_\tau=(u^1_\tau,u^2_\tau,...,u^N_\tau)$ as the concatenated vector of all vehicles' state variables and control inputs at time $\tau$, and furthermore, $U=(u_0,u_1...,u_{T-1})$ represents the concatenated vectors of all vehicles' inputs at all time. Following~\cite{kavuncu2021potential}, the overall cost can be given as
\begin{equation}
J(U)=\sum_{\tau =0}^{T-1}P_\tau (x_\tau,u_\tau) + P_T(x_T)
\end{equation}
where
\begin{align}
P_\tau(x_\tau,u_\tau)&=\sum^N_{i=1}C^i_\tau(x^i_\tau,u^i_\tau )+\sum_{1\leq i<j\leq N}C_\tau^{ij}(x^i_\tau,x^j_\tau ), \\
P_T(x_T) &= \sum_{i=1}^NC^i_T(x^i_T)+\sum_{1\leq i<j\leq N}C^{ij}_T(x^i_T,x^j_T).
\end{align}
With a slight abuse of notation, we use $C^i_\tau$ and $C^i_T$ to denote running cost at time $\tau$ and terminal cost, of vehicle $i$, respectively, while it should be noted that the terminal cost is not a function of control inputs. $C^i_\tau$, the running cost, measures the deviation of vehicle $i$ from its reference position $x^i_{\tau,ref}$ and the control cost at time $\tau$, which is defined as
\begin{equation}
C^i_\tau(x^i_\tau,u^i_\tau)=(x^i_\tau-x^i_{\tau,ref})^\top Q(x^i_\tau-x^i_{\tau,ref})+u^{i\top}_\tau Ru^i_\tau.
\end{equation}
Similarly, the terminal cost $C^i_T$ is defined as
\begin{equation}
C^i_T=(x^i_T-x^i_{T,ref})^\top Q(x^i_T-x^i_{T,ref})
\end{equation}
with $x^i_{T,ref}$ being the reference position of the terminal.
$C^{ij}_\tau$ is the pair-wise collision avoidance cost between vehicle $i$ and vehicle $j$ at time $\tau$, which is defined as
\begin{equation}
C^{ij}_\tau(x^i_\tau,x^j_\tau)=
\left\{
\begin{array}{ll}
\beta(d^{ij}_\tau-d_\textup{safe})^2 & \textup{if}\ d^{ij}_\tau < d_\textup{safe}\\
0 & \textup{else}
\end{array}
\right.
\end{equation}
where $d^{ij}_\tau$ is the Euclidean distance between centers of vehicles $i$ and $j$ at time $\tau$, and $d_\textup{safe}$ is the given safe distance, such that when two vehicles are close to each other, a penalty is added towards the overall cost. $\beta$ is the coefficient that determines the ratio between two types of cost. Theoretically, safety can be guaranteed by picking an excessively large $\beta$ such that the soft penalties perform approximately as hard constraints.

Meanwhile, the control inputs of vehicles are bounded due to various physical limits. Such limits include steering angle limits as well as acceleration limits due to limited engine force and brake force. Therefore, the following box constraints are imposed.
\begin{equation}
\underline{u}^i_\tau\preceq u^i_\tau\preceq \overline{u}^i_\tau, \forall i\in\mathcal{N}, \forall \tau\in\mathcal{T}
\end{equation}

Considering vehicle dynamics, objective functions, and constraints, the following optimization problem can be derived.
\begin{equation}
\label{originalProb}
\begin{array}{lcl}
\min\limits_{U} &\ &J(U)\\
s.t.&\ &x^i_{\tau+1} = f(x^i_\tau,u^i_\tau),\\
&\ &\underline{u}^i_\tau\preceq u^i_\tau\preceq \overline{u}^i_\tau,\\
&\ &\forall i\in\mathcal{N}, \forall \tau\in\mathcal{T}
\end{array}
\end{equation}

\begin{remark}
Such an optimization problem can be viewed as an optimal control problem (OCP) and be solved by the constrained iLQR~\cite{chen2019autonomous}. However, due to the existence of pair-wise collision avoidance costs, solving such a problem in a centralized manner involves the inverse of a dense positive-definite matrix with an element number proportional to $N^2$ at each time stamp, which results in $O(N^3)$ complexity. Such a method suffers from poor scalability with respect to the number of vehicles, and therefore can hardly be applied to traffic scenarios of large scale.
\end{remark}

\section{Decentralized ILQR for Cooperative Trajectory Planningvia Dual Consensus ADMM}
In this section, we perform convex reformulation of the aforementioned OCP (\ref{originalProb}) and present an optimization framework by inheriting the merits of both iLQR and dual consensus ADMM. Such an optimization framework solves the OCP in a fully decentralized and parallelizable manner. As a result, noteworthy scalability with respect to the number of vehicles is exhibited. We assume a complete graph on the communication network structure, which enables each pair of CAVs to exchange information with each other directly.

\subsection{Convex Reformulation}

It should be noted that problem (\ref{originalProb}) is strongly non-convex due to the existence of non-linear vehicle dynamics and the non-convexity brought by collision avoidance costs. Such non-convexity is inherent to most cooperative trajectory planning problems. However, to apply the dual consensus ADMM algorithm, strong duality of the problem is required, which commonly necessitates the convexity. Therefore, we need to reformulate the problem (\ref{originalProb}) as a convex optimization problem in the first place. The convex reformulation includes the linearization of the system dynamics and convex approximation of all the costs around the neighbourhood of nominal vehicle trajectories.

Given a feasible nominal trajectory $\{x^i_\tau,u^i_\tau\}\in \mathbb{R}^M$ for vehicle $i$ where $M=(T+1)n+Tm$ is the overall dimension of the trajectory, linear approximation for vehicle dynamics can be performed as 
\begin{equation}
\begin{aligned}
\delta x^{i}_{\tau+1}&=A^i_\tau\delta x^i_\tau+B^i_\tau \delta u^i_\tau, \\ A^i_\tau&=\frac{\partial f(x^i_\tau,u^i_\tau)}{\partial x^i_\tau},
B^i_\tau=\frac{\partial f(x^i_\tau,u^i_\tau)}{\partial u^i_\tau},
\forall i\in\mathcal{N}.
\end{aligned}
\end{equation}

For all host costs $C^i_\tau$ and terminal cost $C^i_T$, we denote the perturbed costs as $\bar{C}^i_\tau(\delta x^i_\tau, \delta u^i_\tau)=C^i_\tau(x^i_\tau+\delta x^i_\tau, u^i_\tau+\delta u^i_\tau)$ and $\bar{C}^i_T(\delta x^i_T)=C^i_T(x^i_T+\delta x^i_T)$.
By performing Taylor expansion to the second order, we obtain the approximation to such perturbed costs as
\begin{equation}
\begin{aligned}
\bar{C}^i_\tau(\delta x^i_\tau, \delta u^i_\tau)\approx& C^i_\tau (x^i_\tau,u^i_\tau) + \delta u^{i\top}_\tau C^i_{\tau,u} + \delta x^{i\top}_\tau C^i_{\tau,x}\\
&+\frac{1}{2}\delta u^{i\top}_\tau C^i_{\tau,uu}\delta u^{i}_\tau +\frac{1}{2}\delta x^{i\top}_\tau C^i_{\tau,xx}\delta x^{i}_\tau \\
\bar{C}^i_T(\delta x^i_T)\approx &C^i_T(x^i_T)+\delta x^{i\top}_T C^i_{T,x}+\frac{1}{2}\delta x^{i\top}_T C^i_{T,xx}\delta x^{i}_T\\
\end{aligned}
\end{equation}
where $C^i_{\tau,u}$ and $C^i_{\tau,x}$ are the partial derivatives of $C^i_\tau$ with respect to $u^i_\tau$ and $x^i_\tau$, and $C^i_{\tau,uu}$ and $C^i_{\tau,xx}$ are the second-order partial derivatives of $C^i_\tau$ with respect to $u^i_\tau$ and $x^i_\tau$. Similarly, $C^i_{T,x}$ and $C^i_{T,xx}$ are the first-order and second-order derivatives of $C^i_T$ with respect to $x^i_T$, respectively.

We then define the perturbed overall host cost as
\begin{equation}
\label{def_Fhat}
\begin{aligned}
\bar{F}^i(\delta X^i)=&\sum_{\tau=0}^{T-1}\bar{C}^i_\tau(\delta x^i_\tau,\delta u^i_\tau)+\bar{C}^i_T(\delta x^i_T)\\
&-\sum^{T-1}_{\tau=0}C^i_\tau(x^i_\tau,u^i_\tau)-C^i_T(x^i_T)
\end{aligned}
\end{equation}
which can be rewritten in a compact matrix form. We define $\delta X^i=(\delta x^i_0, \delta u^i_0,...,\delta x^i_T)\in\mathbb{R}^M$ as the concatenated vector of variation of all state vectors and input vectors corresponding to vehicle $i$, $L^i_1=(C^i_{0,x},C^i_{0,u},\cdots,C^i_{T,x})\in\mathbb{R}^M$ as the column vector containing all the first-order Jacobians, and $L^i_2=\textup{blocdiag}\{C^{i}_{0,xx},C^{i}_{0,uu},\cdots,C^{i}_{T,xx}\}\in\mathbb{R}^{M\times M}$ as the block diagonal matrix of all second-order Hessians. (\ref{def_Fhat}) can be written as
\begin{equation}
\begin{aligned}
&\bar{F}^i(\delta X^i)= \delta X^{i\top}L^i_1+\frac{1}{2}\delta X^{i\top}L^i_2\delta X^i.
\end{aligned}
\end{equation}

The linearized dynamic constraints can also be expressed in the matrix form as
\begin{equation}
(L^i_3-L^i_4) \delta X^i=0
\end{equation}
where 
$L^i_3=\textup{blocdiag}\{[A^i_0\ B^i_0],[A^i_1\ B^i_1],...,[A^i_{T-2}\ B^i_{T-2}],$
$[A^i_{T-1}\ B^i_{T-1}\ \textbf{0}^{n\times n}]\}$
and
$L^i_4=\textup{blocdiag}\{[\textbf{0}^{n\times n}\ \textbf{0}^{n\times m}\ I^n],$
$\underbrace{[\textbf{0}^{n\times m}\ I^n],...,[\textbf{0}^{n\times m}\ I^n]}\limits_{T-1}\}$.
Such linear equality constraints can be handled by the indicator function and added to the overall costs.

\begin{definition}
The indicator function with respect to a set $\mathbb{\mathcal{X}}$ is defined as
\begin{equation}
\mathcal{I}_\mathcal{X}=
\left\{
\begin{array}{ll}
0 &\textup{if}\ \ x\in\mathcal{X} \\
\infty &\textup{otherwise}
\end{array}
\right.
\end{equation}
\end{definition}
We define $\mathcal{X}^i=\{\delta X^i|(L^i_3-L^i_4)\delta X^i=0\}$, which represents the solution set of the linearized system dynamics. Sum up the indicator function with respect to $\mathcal{X}^i$ with the hosts cost yields
\begin{equation}
\label{FDef}
F^i(\delta X^i):= \bar{F}^i(\delta X^i) + \mathcal I_{\mathcal X^i}(\delta X^i).
\end{equation}
It is obvious that the function $F^i(\delta X^i)$ is convex if and only if all the Hessians in $L^i_2$ are positive semi-definite.

For pairwise collision avoidance terms, using direct Taylor expansion to the second order and calculating the Hessian matrix involves intensive calculation and results in slow performance. Moreover, such an approximation does not produce a convex function in general. Instead, we adopt Gauss-Newton approximation, which not only reduces computation difficulties but also guarantees convexity. We define $\delta x^i=(\delta x^i_0, \delta x^i_1,...,\delta x^i_N)\in\mathbb{R}^{(T+1)n}$, which is obtained by eliminating all control inputs from $\delta X^i$, and
\begin{equation}
l_\tau=[l^{ij}_\tau]_{1\leq i<j\leq N}\in\mathbb{R}^\frac{N(N-1)}{2}, l^{ij}_\tau=\sqrt{\beta}\min\{d^{ij}-d_\textup{safe},0\}
\end{equation}
such that $l_\tau$ is the concatenated column vector of the square root of $C^{ij}_\tau$ for all possible $(i,j)$ pairs satisfying $1\leq i < j \leq N$. With such a definition, the sum of all collision avoidance costs at time $\tau$  can be simply rewritten as the squared $L_2$-norm of $l_\tau$, namely
\begin{equation}
\sum_{1\leq i<j\leq N}C^{ij}_\tau(x^i_\tau,x^j_\tau)=||l_\tau||^2.
\end{equation}
Moreover, the Jacobian of $l_\tau$ with respect to $\delta x^i_\tau$ is defined as $J^i_\tau=\partial l_\tau/\partial x^i_\tau\in\mathbb{R}^{\frac{N(N-1)}{2}\times n}$. 

\begin{remark} 
The Jacobian matrix $J^i_\tau\in\mathbb{R}^{\frac{N(N-1)}{2}\times n}$ contains at most $N-1$ non-zero rows. This is simply due to the fact that only $N-1$ of altogether $N(N-1)/2$ collision avoidance terms are relevant to vehicle $i$. Therefore, $J^i_\tau$ is sparse.
\end{remark}

Furthermore, we define $l=(l_0,l_1,...,l_{T})\in\mathbb{R}^\frac{N(N-1)(T+1)}{2}$. With all the above definitions, we can perform a linear approximation of $l_\tau$ around the current nominal trajectories and obtain the Gauss-Newton approximation of collision avoidance costs. We define the perturbed collision avoidance cost $\bar{C}^{ij}_\tau(\delta x^i_\tau, \delta x^j_\tau)=C^{ij}_\tau(x^i_\tau+\delta x^i_\tau, x^j_\tau+\delta x^j_\tau)$, such that
\begin{equation}
\label{GN_Cij}
\sum_{1\leq i<j\leq N}\bar{C}^{ij}_\tau(\delta x^i_\tau, \delta x^j_\tau)\approx\left\|\sum_{i=0}^NJ_\tau^i\delta x_\tau^i+l_\tau\right\|^2
\end{equation}
Summing (\ref{GN_Cij}) over all time stamps gives
\begin{equation}
\begin{aligned}
&\sum_{\tau=0}^T\sum_{1\leq i<j\leq N}\bar{C}^{ij}_\tau(\delta x^i_\tau, \delta x^j_\tau)\approx\sum_{\tau=0}^T\left\|\sum_{i=0}^NJ_\tau^i\delta x_\tau^i+l_\tau\right\|^2 \\
&=\left\|(\sum_{i=0}^NJ^i_0\delta x^i_0,\sum_{i=0}^NJ^i_1\delta x^i_1,...,\sum_{i=0}^NJ^i_T\delta x^i_T)+l\right\|^2\\
&=\left\|\sum_{i=0}^N(J^i_0\delta x^i_0,J^i_1\delta x^i_1,...,J^i_T\delta x^i_T)+l\right\|^2\\
&=\left\|\sum^N_{i=1}\textup{blocdiag}\{J^i_0,J^i_1,...,J^i_T\}\delta x^i+l\right\|^2.
\end{aligned}
\end{equation}

To also include control inputs in the expression, we define \\
$\tilde{J}^i=\textup{blocdiag}\{[J^i_0\ \textbf{0}^{\frac{N(N-1)}{2}\times m}],...,[J^i_{T-1}\ \textbf{0}^{\frac{N(N-1)}{2}\times m}],J^i_T\}$, which yields
\begin{equation}
\label{sumCollisionAvoid}
\sum_{\tau=0}^N\sum_{1\leq i<j\leq N}\bar{C}^{ij}_\tau(\delta x^i_\tau, \delta x^j_\tau)\approx G_1\left(\sum_{i=1}^N \tilde{J}^i\delta X^i\right)
\end{equation}
where
\begin{equation}
\label{GDef}
G_1(x):=||x+l||^2
\end{equation}
is a convex function. 

Moreover, to include the box constraints imposed on control inputs, we define the convex set $\mathcal{X}^u=\{u_b|\underline{u}-u\preceq u_b\preceq\overline{u}-u\}$, where $\underline{u}=((\underline{u}^1_0,...,\underline{u}^1_{T-1}),...,(\underline{u}^N_0,...,\underline{u}^N_{T-1}))$ and
$\overline{u}=((\overline{u}^1_0,...,\overline{u}^1_{T-1}),...,(\overline{u}^N_0,...,\overline{u}^N_{T-1}))$
denotes the lower bounds and the upper bounds of all the control inputs, and $u=((u^1_0,...,u^1_{T-1}),...,(u^N_0,...,u^N_{T-1}))$ denotes all control inputs in the current nominal trajectories for vehicles. It is obvious that $\delta u=((\delta u^1_0,...,\delta u^1_{T-1}),...,(\delta u^N_0,...,\delta u^N_{T-1}))$ must satisfy $\delta u\in\mathcal{X}^u$ so that the updated control inputs $u+\delta u$ satisfies the box constraints. 

Define a matrix
$O^i=\textup{blocdiag}\{\underbrace{[\textbf{0}^{m\times n} I^m],...,[\textbf{0}^{m\times n} I^m]}\limits_T,$ \\
$\textbf{0}^{m\times n}\}\in\mathbb{Z}^{Tm\times M}$
such that $\delta u^i=O^i\delta X^i$, and furthermore \\
$\tilde{O}^i=[\underbrace{\textbf{0}^{M\times Tm},...,\textbf{0}^{M\times Tm}}\limits_{i-1},O^{i\top},\underbrace{\textbf{0}^{M\times Tm},...,\textbf{0}^{M\times Tm}}\limits_{N-i}]^\top$
$\in\mathbb{R}^{NTm\times M}$
such that $\delta u=\sum_{i=1}^N\tilde{O}^i\delta X^i$. All the box constraints can therefore be handled by the following convex indicator function
\begin{equation}
G_2(\delta u):=\mathcal{I}_{\mathcal{X}^u}(\delta u)=\mathcal{I}_{\mathcal{X}^u}\left(\sum_{i=1}^N\tilde{O}^i\delta X^i\right)
\end{equation}
such that when $u+\delta u$ violates the box constraints, an infinite cost is added to the overall cost. 

We then show that the box constraint cost can be added to the collision avoidance cost with the same form as (\ref{sumCollisionAvoid}) still maintained. Define a new matrix $J^i=[\tilde{J}^{i\top}\tilde{O}^{i\top}]^\top$, and a new convex function
\begin{equation}
G(x)=G\left(\begin{bmatrix}x_{[1]} \\ x_{[2]}\end{bmatrix}\right):=G_1(x_{[1]})+G_2(x_{[2]})
\end{equation}
where $x\in\mathbb{R}^{\frac{1}{2}N(N-1)(T+1)+NTm}$, and $x_{[1]}$, $x_{[2]}$ refer to the first $\frac{1}{2}N(N-1)(T+1)$ elements and the last $NTm$ elements of $x$, respectively. Then
\begin{equation}
\begin{aligned}
G\left(\sum_{i=1}^NJ^i\delta X^i\right)&=G\left(\begin{bmatrix}
\sum_{i=1}^N\tilde{J}^i\delta X^i \\
\sum_{i=1}^N\tilde{O}^i\delta X^i
\end{bmatrix}\right)\\
&=G_1\left(\sum_{i=1}^N\tilde{J}^i\delta X^i\right)+G_2\left(\sum_{i=1}^N\tilde{O}^i\delta X^i\right)
\end{aligned}
\end{equation}
represents the sum of all collision avoidance costs as well as the indicator function that enforce the box constraints. Add it to the host costs and we have the following unconstrained optimization problem
\begin{equation}
\label{NewProb}
\min_{\delta X^1,...,\delta X^N} \sum_{i=1}^NF^i(\delta X^i)+G\left(\sum_{i=1}^N J^i\delta X^i\right)
\end{equation}
which is the convex reformulation of (\ref{originalProb}).

\subsection{Decentralized and Parallelizable Optimization Framework}
In this section, we propose a novel decentralized and parallelizable optimization framework based on dual consensus ADMM to solve (\ref{originalProb}). The following preliminaries concerning dual consensus ADMM need to be restated for completeness (refer to~\cite{banjac2019decentralized,grontas2022distributed} for details).

The convex optimization problem (\ref{NewProb}) can be rewritten as
\begin{equation}
\label{ConstrainedNewProb}
\begin{aligned}
\min_{w,\delta X^1,...,\delta X^N}&\ \ \sum^N_{i=1}F^i(\delta X^i)+G(w) \\
s.t.&\ \ \sum^N_{i=1}J^i \delta X^i=w.
\end{aligned}
\end{equation}
For this problem, slater's condition holds given affine constraints, and therefore strong duality follows.

The Lagrangian of (\ref{ConstrainedNewProb}) is
\begin{equation}
\mathcal{L}(\delta X,w,y)=\sum^N_{i=1}F^i(\delta X^i)+G(w)+y^\top\left(\sum^N_{i=1}J^i\delta X^i-w\right)
\end{equation}
where $y\in\mathbb{R}^{\frac{1}{2}N(N-1)(T+1)+NTm}$ is the dual variable. The dual function can then be given as
\begin{equation}
h(y)=\inf_{\delta X,w}\mathcal{L}(\delta X,w,y)=-\sum^N_{i=1}F^{i*}(-J^{i\top} y)-G^*(y)
\end{equation}
where $F^{i*}$ and $G^*$ are the conjugates of $F^i$ and $G$. The dual problem $\max_{y}h(y)$ is a decomposed consensus optimization problem with all agents making the joint decision on a common optimization variable $y$. Therefore, consensus ADMM~\cite[Alg. 1]{grontas2022distributed} is applicable with $\theta_i:=F^*_i\circ(-J^{i\top})$ and $\xi_i:=\frac{1}{N}G^*$.

\begin{algorithm}[H]
\caption{Consensus ADMM~\cite{grontas2022distributed}}\label{alg:alg1}
\begin{algorithmic}[1]
\State \textbf{choose} $\sigma,\rho >0$
\State \textbf{initialize} for all $i \in \mathcal{N}$: $p^{i,0}=y^{i,0}=z^{i,0}=s^{i,0}=0$
\State \textbf{repeat}: for all $i \in \mathcal{N}$
\State \hspace{0.5cm} Broadcast $y^{i,k}$ to all other vehicles
\State \hspace{0.5cm} $p^{i,k+1}=p^{i,k}+\rho\sum_{j\neq i}(y^{i,k}-y^{j,k})$
\State \hspace{0.5cm} $s^{i,k+1}=s^{i,k}+\sigma(y^{i,k}-z^{i,k})$
\State \hspace{0.5cm} $y^{i,k+1}=\arg\min_{y^i}\{\theta^i(y^i)+y^{i\top}(p^{i,k+1}+s^{i,k+1})$
\Statex \hspace{1.5cm} $+\frac{\sigma}{2}||y^i-z^{i,k}||^2+\rho\sum_{j\neq i}||y^i-\frac{y^{i,k}+y^{j,k}}{2}||^2\}$
\State \hspace{0.5cm} $z^{i,k+1}=\arg\min_{z^i}\{\xi^i(z^i)-z^{i\top}s^{i,k+1}$
\Statex \hspace{1.5cm} $+\frac{\sigma}{2}||z^i-y^{i,k+1}||^2\}$
\State \textbf{until} termination criterion is satisfied
\end{algorithmic}
\label{alg1}
\end{algorithm}

In particular, Step 7 of Algorithm 1 is performed by
\begin{equation}
\label{yUpdate}
y^{i,k+1}=\frac{1}{\sigma+2\rho d_i}(J^i \delta X^{i,k+1}+r^{i,k+1}),
\end{equation}
where $r^{i,k+1}=\rho\sum_{j\neq i}(y^{i,k}+y^{j,k})+\sigma z^{i,k}-p^{i,k+1}-s^{i,k+1}.$
\begin{equation}
\label{xUpdate}
\delta X^{i,k+1}=\underset{{\delta X^i}}{\arg\min}\{F^i(\delta X^i)+\frac{||J^i \delta X^i+r^{i,k+1}||^2}{2(\sigma+2\rho d_i)}\}
\end{equation} 
is an auxiliary variable and converges to a minimizer of (\ref{NewProb})~\cite{grontas2022distributed}. $d_i$ is the degree of node $i$ which equals to $N-1$ in our case. Meanwhile, Step 8 of Algorithm 1 is performed by
\begin{equation}
\label{zUpdate0}
z^{i,k+1}=\frac{s^{i,k+1}}{\sigma}+y^{i,k+1}-\frac{1}{N\sigma}\textup{Prox}^{\frac{1}{N\sigma}}_G(N(s^{i,k+1}+\sigma y^{i,k+1})).
\end{equation}

Based on the previous restatement of the existing dual consensus ADMM method, we propose the following dual update and the primal solution recovery method that are specific to our optimization problem. In particular, (\ref{xUpdate}) and (\ref{zUpdate0}) require further discussion.

\subsubsection{Dual Update}
In (\ref{zUpdate0}), the first $\frac{1}{2}N(N-1)(T+1)$ elements and the last $NTm$ elements of $z^{i,k+1}$ correspond to collision avoidance costs and box constraints, respectively, and therefore they should be handled separately. The last term is
\begin{equation}
\begin{aligned}
&\textup{Prox}^{\frac{1}{N\sigma}}_G(N(s^{i,k+1}+\sigma y^{i,k+1}))\\
&=\underset{z}{\arg\min}\{G(z)+\frac{1}{2N\sigma}||z-N(s^{i,k+1}+\sigma y^{i,k+1})||^2\}\\
&=(\underset{z_{[1]}}{\arg\min}\{||z_{[1]}+l||^2\\
&+\frac{1}{2N\sigma}||z_{[1]}-N(s^{i,k+1}_{[1]}+\sigma y^{i,k+1}_{[1]})||^2\},\\
&\underset{z_{[2]}}{\arg\min}\{I_{\mathcal{X}^u}(z_{[2]})+\frac{1}{2N\sigma}||z_{[2]}-N(s^{i,k+1}_{[2]}+\sigma y^{i,k+1}_{[2]})||^2\}).
\end{aligned}
\end{equation}
Thus, we decompose the problem into two parallel sub-problems. The first sub-problem is a simple unconstrained quadratic program, and the analytical solution is given as
\begin{equation}
z_{[1]}=\frac{N}{2N\sigma+1}(s_{[1]}^{i,k+1}+\sigma y_{[1]}^{i,k+1}-2\sigma l).
\end{equation}
Plug it back to (\ref{zUpdate0}) and we have
\begin{equation}
\label{zUpdate2}
z^{i,k+1}_{[1]}=\frac{2}{2N\sigma+1}(Ns_{[1]}^{i,k+1}+N\sigma y_{[1]}^{i,k+1}+l)
\end{equation}
which performs weighted sum of $s$, $y$, and $l$.

The second sub-problem is equivalent to the following constrained optimization problem:
\begin{equation}
\label{zUpdate2sub}
\begin{aligned}
\min_{z_{[2]}}&\ ||z_{[2]}-N(s^{i,k+1}_{[2]}+\sigma y^{i,k+1}_{[2]})||^2 \\
s.t.&\ z_{[2]}\in\mathcal{X}^u,
\end{aligned}
\end{equation}
and $z_{[2]}$ is a minimizer of (\ref{zUpdate2sub}). To solve for $z_{[2]}$, we first introduce the following definition of projection.
\begin{definition}
Given a closed convex set $\mathcal{X}$ in $\mathcal{D}$, there exists a unique minimizer for every $v\in\mathcal{D}$ to the following problem
$\min_x\{||x-v||^2|x\in\mathcal{X}\}$
, which is called projection of $v$ onto $\mathcal{X}$ and denoted as $\textup{Proj}_\mathcal{X}(v)$.
\end{definition}

From this definition, the solution to the second sub-problem is
\begin{equation}
\label{z2Update2}
z^{i,k+1,*}_{[2]}=\textup{Proj}_{\mathcal{X}^u}(N(s^{i,k+1}_{[2]}+\sigma y^{i,k+1}_{[2]})).
\end{equation}
Since the convex set $\mathcal{X}^u$ is constructed by imposing separate box constraints on each element of the vector $u_b$, the result of this projection is simply given by confining each element of $N(s^{i,k+1}_{[2]}+\sigma y^{i,k+1}_{[2]})$ into the set of $\mathcal{X}^u$ respectively, which can be performed by element-wise min-max operation. Plug it back to (\ref{zUpdate0}) and we have
\begin{equation}
\label{z2Update3}
z^{i,k+1}_{[2]} = \frac{s^{i,k+1}}{\sigma}+y^{i,k+1}-\frac{1}{N\sigma}z^{i,k+1,*}_{[2]}.
\end{equation}
Finally, we have $z^{i,k+1}=(z^{i,k+1}_{[1]},z^{i,k+1}_{[2]})$.

Such results reveal that $z$ is updated by element-wise addition and min-max operation, and therefore can be performed efficiently. Equations (\ref{zUpdate2}), (\ref{z2Update2}), (\ref{z2Update3}), and (\ref{yUpdate}) form the steps of update for dual variables $y$ and $z$.

\subsubsection{Primal Solution Recovery}
Perform expansion on the square term of (\ref{xUpdate}) yields
\begin{equation}
\label{xUpdate2}
\begin{aligned}
&\delta X^{i,k+1}=\underset{{\delta X^i}}{\arg\min}\{F^i(\delta X^i)\\
&+\frac{1}{2(\sigma +2\rho d_i)}(\delta X^{i\top}J^{i\top}J^i\delta X^i+2r^{i,k+1\top}J^i\delta X^i)\}
\end{aligned}
\end{equation}
Utilizing the sparse structure of the Jacobian matrix $J^i=[\tilde{J}^\top \tilde{O}^\top]^\top$, we can obtain the following result for second-order terms of $\delta X^i$
\begin{equation}
\label{expansion1}
\begin{aligned}
\delta X^{i\top}J^{i\top}J^i\delta X^i&=\delta X^{i\top}\tilde{J}^{i\top}\tilde{J}^i\delta X^i+\delta X^{i\top}\tilde{O}^{i\top}\tilde{O}^i\delta X^i\\
&=\delta X^{i\top}\tilde{J}^{i\top}\tilde{J}^i\delta X^i+\delta u^{i\top}\delta u^i\\
&=\sum_{\tau=0}^T\delta x^{i\top}_\tau J^{i\top}_\tau J^i_\tau\delta x^i_\tau +\sum_{\tau=0}^{T-1}\delta u^{i\top}_\tau\delta u^i_\tau.
\end{aligned}
\end{equation}
Similarly, for first-order terms of $\delta X^i$, we have
\begin{equation}
\label{expansion2}
\begin{aligned}
2r^{i,k+1\top}J^i\delta X^i&=2r^{i,k+1\top}_{[1]}\tilde{J}^i\delta X+2r^{i,k+1\top}_{[2]}\tilde{O}^i\delta X\\
&=2r^{i,k+1\top}_{[1]}\tilde{J}^i\delta X+2\tilde{r}^{i,k+1\top}_{[2]}\delta u^i\\
&=\sum_{\tau=0}^T2r^{i,k+1\top}_{[1],\tau} J^{i}_\tau \delta x^i_\tau +\sum_{\tau=0}^{T-1}2\tilde{r}^{i,k+1\top}_{[2],\tau} \delta u^i_\tau.
\end{aligned}
\end{equation}
In particular, $r_{[1],\tau}$ represents the piece of $r_{[1]}$ from row number $N(N-1)\tau/2$ to row number $N(N-1)(\tau+1)/2-1$, $\tilde{r}^i_{[2]}$ represents the piece of $r^i_{[2]}$ from row number $(i-1)Tm$ to row number $iTm-1$, and $\tilde{r}^i_{[2],\tau}$ represents the piece of $\tilde{r}^i_{[2]}$ from row number $\tau m$ to $(\tau+1)m-1$, inclusively. 

Plug (\ref{FDef}), (\ref{expansion1}), and (\ref{expansion2}) back into (\ref{xUpdate2}), and we can show that $\delta X^{i,k+1}$ is the optimizer of the following minimization problem
\begin{equation}
\label{LQR}
\begin{aligned}
\min_{\delta x^i_0, \delta u^i_0,...,\delta x^i_T} &\ 
\sum_{\tau=0}^T \frac{1}{2}\delta x^{i\top}_\tau(C^i_{\tau,xx}+\frac{J^{i\top}_\tau J^i_\tau}{2(\sigma+2\rho d_i)})\delta x^i_\tau \\
&+\delta x^{i\top}_\tau(C^i_{\tau,x}+\frac{J^{i\top}_\tau r^{i,k+1}_{[1],\tau}}{\sigma+2\rho d_i})\\
&+\sum_{\tau=0}^{T-1}\frac{1}{2}\delta u^{i\top}_\tau(C^i_{\tau,uu}+\frac{I}{2(\sigma+2\rho d_i)})\delta u^i_\tau \\
&+\delta u^{i\top}_\tau(C^i_{\tau,u}+\frac{\tilde{r}^{i,k+1}_{[2],\tau}}{\sigma+2\rho d_i})
\\
s.t. &\	\delta x^i_{\tau+1}=A^i_\tau \delta x^i_\tau+B^i_\tau \delta u^i_\tau.
\end{aligned}
\end{equation}
Remarkably, (\ref{LQR}) is a standard LQR optimal control problem and can be solved efficiently via dynamic programming.
\begin{remark}
With the introduction of problem (\ref{LQR}), the originally coupled optimization problem is essentially decoupled, as each vehicle is solving an optimal control problem that only involves its own state and input variables. The influence of other vehicles on their own trajectory is reflected by the additional terms containing $J$ and $r$. Iteratively solving the LQR problem (\ref{LQR}) by each vehicle until ADMM termination corresponds to a single backward pass in the centralized iLQR solver, where state variables of all vehicles are involved. Based on the iterative convex reformulation of the original problem around current nominal trajectories, a decentralized iLQR algorithm is introduced.
\end{remark}

Based on the above discussion, we propose Algorithm 2, which is a decentralized version of the iLQR algorithm that solves problem (\ref{originalProb}). Each vehicle performs two loops: the outer loop performs convex reformulation of problem (\ref{originalProb}), which resembles the outer loop of a centralized iLQR solver. The inner loop performs ADMM iterations to solve the introduced convex problem in a decentralized and parallelizable manner, during which each vehicle solves an LQR problem (\ref{LQR}) as well as performing updates of dual variables $y$, $z$, $p$, and $s$. 

\begin{algorithm}[t]
\caption{Decentralized iLQR via Dual Consensus ADMM (for vehicle $i$)}\label{alg:alg1}
\begin{algorithmic}[1]
\State \textbf{initialize} $\{x^i_\tau,u^i_\tau\}^T_{\tau=0}$
\State \textbf{initialize} for all $i \in \mathcal{N}$: 
\Statex \hspace{0.5cm} $p^{i,0}=y^{i,0}=z^{i,0}=s^{i,0}=0$
\State \textbf{choose} $\sigma,\rho >0$
\State \textbf{repeat}:
\State \hspace{0.5cm} Send $\{x^i_\tau\}_{\tau=1}^T$, receive $\{x^j_\tau\}_{\tau=1}^T$ from $j\in\mathcal{N}-\{i\}$
\State \hspace{0.5cm} Compute $l$, $J^i$, $\{A^i_\tau\}^{T-1}_{\tau=0}$, $\{B^i_\tau\}^{T-1}_{\tau=0}$
\State \hspace{0.5cm} \textbf{reset} $p^{i,0}=s^{i,0}=0$
\State \hspace{0.5cm} \textbf{repeat}:
\State \hspace{1.0cm} Send $y^{i,k}$, receive $y^{j,k}$ from $j\in\mathcal{N}-\{i\}$
\State \hspace{1.0cm} $p^{i,k+1}=p^{i,k}+\rho\sum_{j\neq i}(y^{i,k}-y^{j,k})$
\State \hspace{1.0cm} $s^{i,k+1}=s^{i,k}+\sigma(y^{i,k}-z^{i,k})$
\State \hspace{1.0cm} $r^{i,k+1}=\rho\sum_{j\neq i}(y^{i,k}+y^{j,k})$
\Statex \hspace{2.0cm} $+\sigma z^{i,k}-p^{i,k+1}-s^{i,k+1}$
\State \hspace{1.0cm} Compute $\delta X^{i,k+1}$ by solving LQR problem (\ref{LQR})
\State \hspace{1.0cm} Update $y^{i,k+1}$ using (\ref{yUpdate})
\State \hspace{1.0cm} Update $z^{i,k+1}$ using (\ref{zUpdate2}) and (\ref{z2Update2})
\State \hspace{0.5cm} \textbf{until} termination criterion is satisfied
\State \hspace{0.5cm} Update $\{x^i_\tau,u^i_\tau\}^T_{\tau=0}$
\State \textbf{until} termination criterion is satisfied
\end{algorithmic}
\label{alg2}
\end{algorithm}
\begin{remark}
\label{initialization}
It is important to notice that during Step 7 of Algorithm 2, only $p$ and $s$ are reset. The values of $y$ and $z$ obtained by the previous ADMM iterations are carried through to the next ADMM iterations and act as initialization. The principle behind this is that the induced convex optimization problem does not change a lot between consecutive convex reformulations: it is only drifting slowly. Therefore, the results of dual variables $y$ and $z$ from the previous convex reformulation loop serve as a good guess to the optimal $y^*$ and $z^*$ for the next loop, thus fastening the ADMM convergence greatly. 
\end{remark}

\subsection{Dynamically Feasible Trajectory Update}
It should be noted that $\delta X^i$ for vehicle $i$ obtained through the solving of problem (\ref{LQR}) is only feasible for the linearized system dynamics but not feasible for the original nonlinear system dynamics. Therefore, the direct addition of $\delta X^i$ and the nominal trajectory causes a violation of the nonlinear dynamic constraints. To handle this problem, we propose a dynamically feasible update strategy, which corresponds to Step 17 of Algorithm 2.

Solving problem (\ref{LQR}) via dynamic programming yields a series of feedback control matrices $\{k_\tau^i, K_\tau^i\}_{\tau=0}^{T-1}$ such that
\begin{equation}
\delta u_\tau^i=k_\tau^i+K^i_\tau\delta x^i_\tau, \forall \tau \in \mathcal{T}.
\end{equation}

Following the typical iLQR forward pass, we update the nominal trajectory as
\begin{equation}
\label{trueUpdate}
\begin{aligned}
u^i_\tau &= \hat{u}^i_\tau+\alpha k^i_\tau+K^i_\tau(x^i_\tau-\hat{x}^i_\tau)\\
u^i_\tau &\leftarrow \textup{clip}(u^i_\tau) \\
x^i_{\tau+1} &= f(x^i_\tau,u^i_\tau),
\end{aligned}
\end{equation}
where $\alpha$ is the line search parameter. Clipping of inputs on each time stamp is performed such that the inputs satisfy the box constraints strictly.

Based on the previous discussion, we propose Algorithm 3 to perform dynamically feasible trajectory updates for each vehicle. For better convergence, line search method is used. In Algorithm 3, each vehicle iterates the line search parameter $\alpha$ through the same list of candidate $\alpha$ and updates its trajectory by (\ref{trueUpdate}) to generate a set of candidate trajectories. After that, the candidate trajectories are broadcast to all other vehicles. Then, each vehicle finds the trajectory corresponding to the optimal $\alpha$ that produces the lowest overall cost, and uses it to update the current trajectory. Noted that in Algorithm 3, synchronization only takes place at Step 5, which causes a minor impact on the overall algorithm.
\begin{algorithm}[t]
\caption{Dynamically Feasible Update with Line Search (for vehicle $i$)}\label{alg:alg1}
\begin{algorithmic}[1]
\State \textbf{initialize} $t\_list^i,c\_list^i$ = $empty$
\State \textbf{for} $\alpha$ \textbf{in} $\alpha\_list$:
\State \hspace{0.5cm} Compute $traj$ with $\alpha$ using (\ref{trueUpdate})
\State \hspace{0.5cm} Append $traj$ to the $t\_list^i$
\State Send $t\_list^i$, receive $t\_list^j$ from $j\in\mathcal{N}-\{i\}$
\State \textbf{for} i \textbf{in} $len(\alpha\_list)$:
\State \hspace{0.5cm} $t\_set$ = $empty$
\State \hspace{0.5cm} \textbf{for} j \textbf{in} $\mathcal{N}$:
\State \hspace{1.0cm} Append $t\_list^j[i]$ to $t\_set$
\State \hspace{0.5cm} Compute $cost$ of $t\_set$
\State \hspace{0.5cm} Append $cost$ to $c\_list$
\State $index$ = $\arg\min(c\_list)$
\State $current\ trajectory$ $\leftarrow$ $t\_list^i[index]$
\end{algorithmic}
\label{alg2}
\end{algorithm}

% To this point, the overview of Algorithms 2 and 3 is shown in Fig. 1.

\subsection{Analysis of Complexity}
Here, we give a brief discussion on the complexity of the proposed algorithm Algorithm 2. In our simulations, the most computationally demanding part of Algorithm 2 is Step 13, which requires solving an LQR optimal control problem. It is obvious that for each vehicle, problem (\ref{LQR}) involves only state variables and control inputs local to that vehicle, and therefore its scale does not grow with the number of CAVs. In other words, we can conclude that solving problem (\ref{LQR}) is of $O(1)$ complexity.

For comparison, solving problem (\ref{originalProb}) in a centralized manner tackles an LQR optimal control problem with the state space growing linearly with respect to the number of CAVs, which results in poor scalability. Generally, $O(N^3)$ complexity is induced.

It should be noted that problem (\ref{LQR}) needs to be solved by each vehicle in an iterative manner. With the assumption that the number of iterations for the outer loop is $N_1$ and the number of iterations for the inner loop is $N_2$, we conclude that if only Step 13 is considered, the complexity of Algorithm 2 is $O(N_1\cdot N_2)$.

Other steps such as Steps 10 and 12 have higher complexity theoretically, but they only involve very simple operations such as element-wise summation, and therefore only take up a negligible amount of time in our simulations.

% \begin{figure}[t]
% \label{fig:Overview}
% \includegraphics[scale=0.32]{pictures/Flowchart.png}
% \caption{Overview of the decentralized and parallelizable optimization framework and dynamically feasible trajectory update strategy.}
% \end{figure}

\section{Simulation Results}
\subsection{Vehicle Model}
We assume that all vehicles involved in our simulation possess the same dynamics. Referring to~\cite{tassa2014control}, Single vehicle dynamics is characterized by the following equations:
\begin{equation}
\label{dynamics}
\begin{aligned}
p_{x,\tau+1} &= p_{x,\tau}+f_r(v_\tau,\delta_\tau)\cos(\theta_\tau),\\
p_{y,\tau+1} &= p_{y,\tau}+f_r(v_\tau,\delta_\tau)\sin(\theta_\tau),\\
\theta_{\tau+1} &= \theta_{\tau}+\arcsin(\frac{\tau_s v_\tau\sin(\delta_\tau)}{b}),\\
v_{\tau+1} &= v_\tau+\tau_sa_\tau.
\end{aligned}
\end{equation}
(\ref{dynamics}) describes a discrete-time vehicle model with state vector $(p_x,p_y,\theta,v)$ and input vector $(\delta,a)$. $p_x$ and $p_y$ represent the global X and Y Cartesian coordinates of the vehicle center, $\theta$ is the heading angle of the vehicle with respect to the positive X axis of the global Cartesian coordinate system, $v$ is the velocity of the vehicle, $\delta$ is the steering angle of the front wheel, and $a$ is the acceleration. Moreover, we use the subscript $[\cdot]_\tau$ to indicate variables of time stamp $\tau$ with $\tau\in\mathcal{T}$, and $\tau_s$ is the time interval. The function $f_r(v,\delta)$ is defined as
\begin{equation}
f_r(v,\delta) = b+\tau_sv\cos(\delta)-\sqrt{b^2-(\tau_sv\sin(\delta))^2}
\end{equation}
where $b$ denotes the vehicle wheelbase.

The system dynamics of all vehicles considered as a single system can be obtained by trivially stacking all single vehicle dynamics together, yielding a model with $4N$ state variables and $2N$ control inputs.

\begin{figure*}[h]
\centering
\subfigure[$\tau=0$]{\includegraphics[scale=0.30]{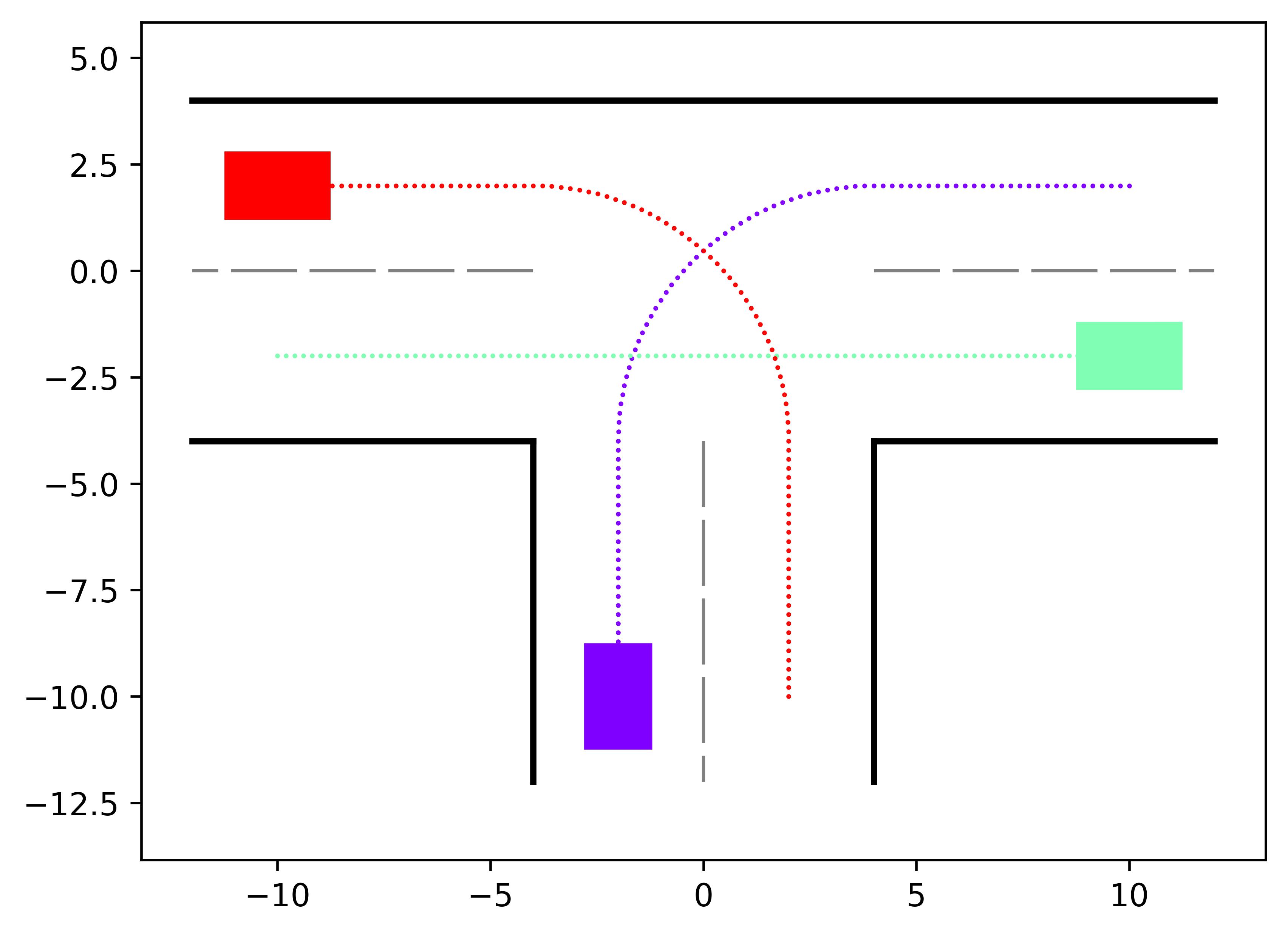}}
\subfigure[$\tau=50$]{\includegraphics[scale=0.30]{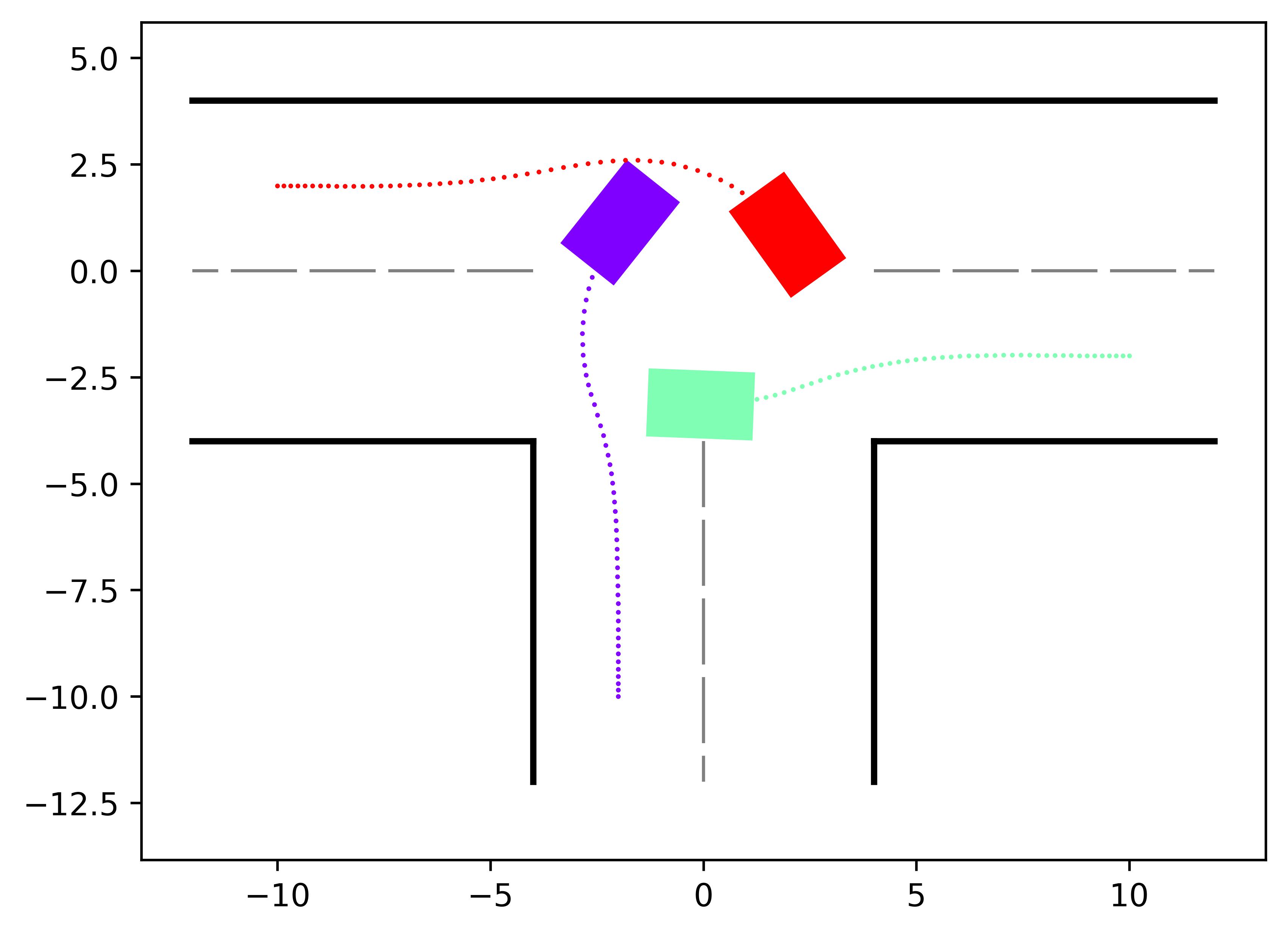}}
\subfigure[$\tau=75$]{\includegraphics[scale=0.30]{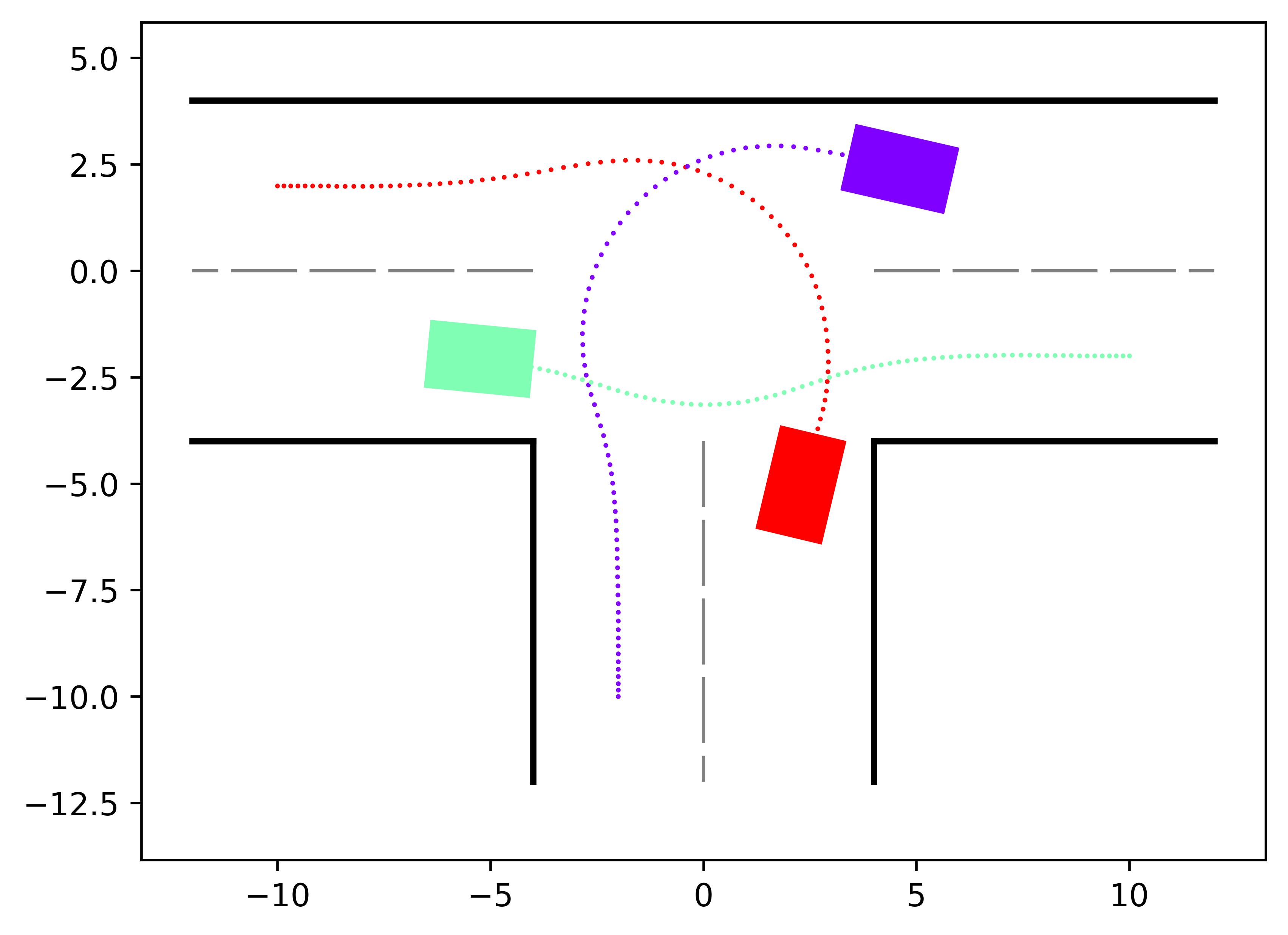}}
\subfigure[$\tau=99$]{\includegraphics[scale=0.30]{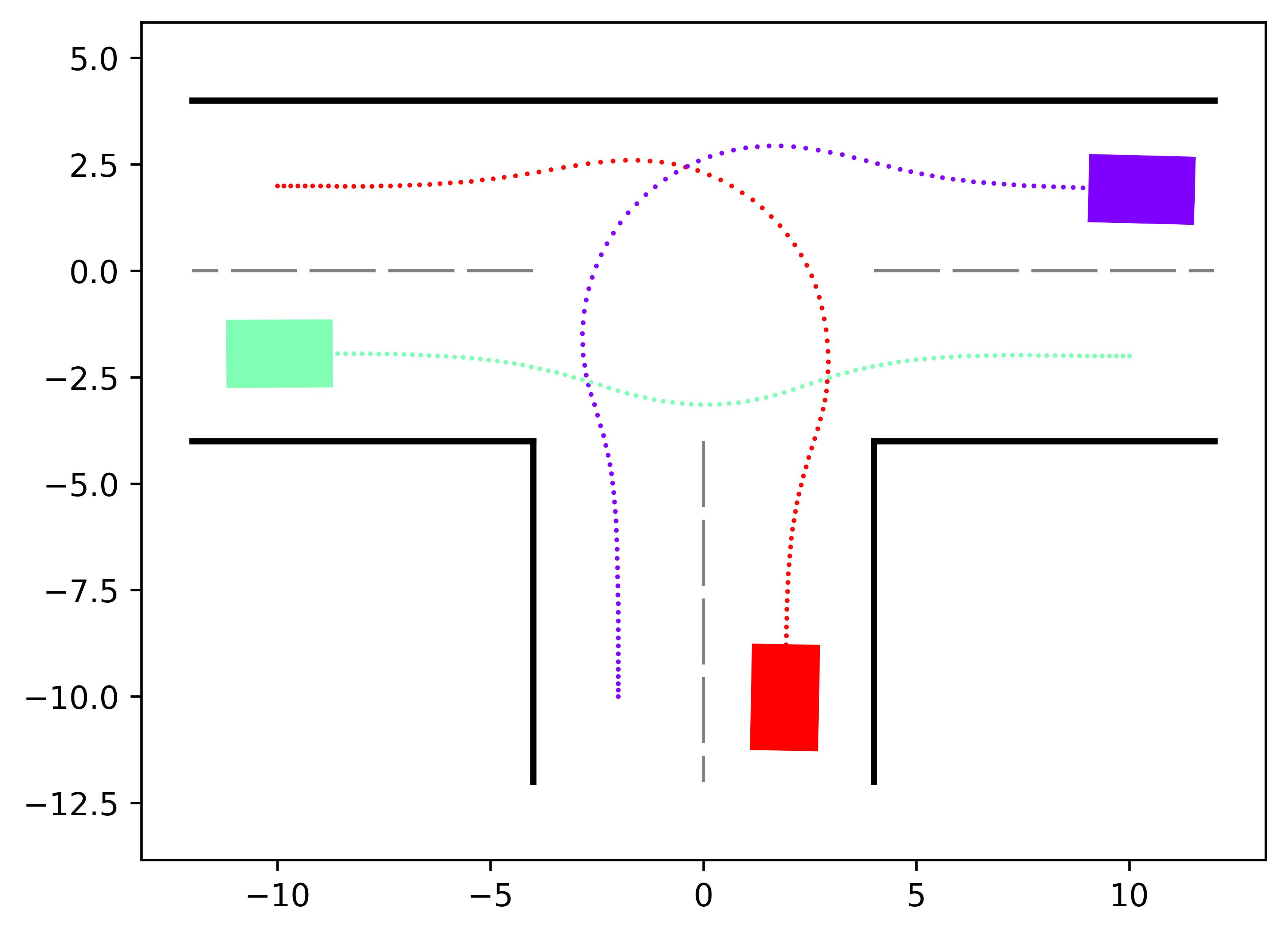}}
\caption{Simulation results for Scenario 1 with the proposed method at time stamps $\tau=0$, $\tau=50$, $\tau=75$, and $\tau=99$.}
\label{fig:scenario1Results}
\end{figure*}

\subsection{General Settings}
In this paper, we consider the same scenarios as in~\cite{zhang2021semi}, which includes a T-junction with 3 vehicles and an intersection with 12 vehicles. We implement the proposed algorithm with both a single-process version and a multi-process version with the number of processes equals to the number of vehicles. We also implement three baselines for comparison, including a centralized iLQR solver with log barrier functions~\cite{chen2019autonomous}, an IPOPT solver, and an SQP solver, with the last two solvers provided by CasADi~\cite{andersson2019casadi}. All algorithms are implemented in Python 3.7, running on a server with $2 \times$ Intel(R) Xeon(R) Gold 6348 CPU @ 2.60GHz. For multi-process implementation, each process is bound to a (logical) core on the CPU, and the communication is realized via shared memory.

For details, we set the length and width of each vehicle as $2.50\,\textup{m}$ and $1.60\,\textup{m}$ respectively. The input steering angle is bounded between $\pm0.6\,\textup{rad}$ and the acceleration is within $[-3.0,1.5]\,\textup{m}/\textup{s}^2$. We set the weighting matrices in the host cost as $Q=\textup{diag}(1,1,0,0)$ and $R=\textup{diag}(1,1)$. The time interval for discrete dynamics is set to be $\tau_s=0.1\,\textup{s}$ and the horizon length is $T=100$. The safe distance for all collision avoidance costs is $d_\textup{safe}=5.5\,\textup{m}$, and the scaling factor between host cost and collision avoidance cost is $\beta=1.44$. Trajectories of all vehicles are initialized with zero inputs. The terminal condition of the proposed method and the centralized iLQR solver is set according to the change of overall cost. When the absolute change of overall cost between two consecutive iterations is less than 1, both algorithms terminate.

\subsection{Main Results}
\subsubsection{Scenario 1}

\begin{figure}[t]
\centering
\includegraphics[scale=0.50]{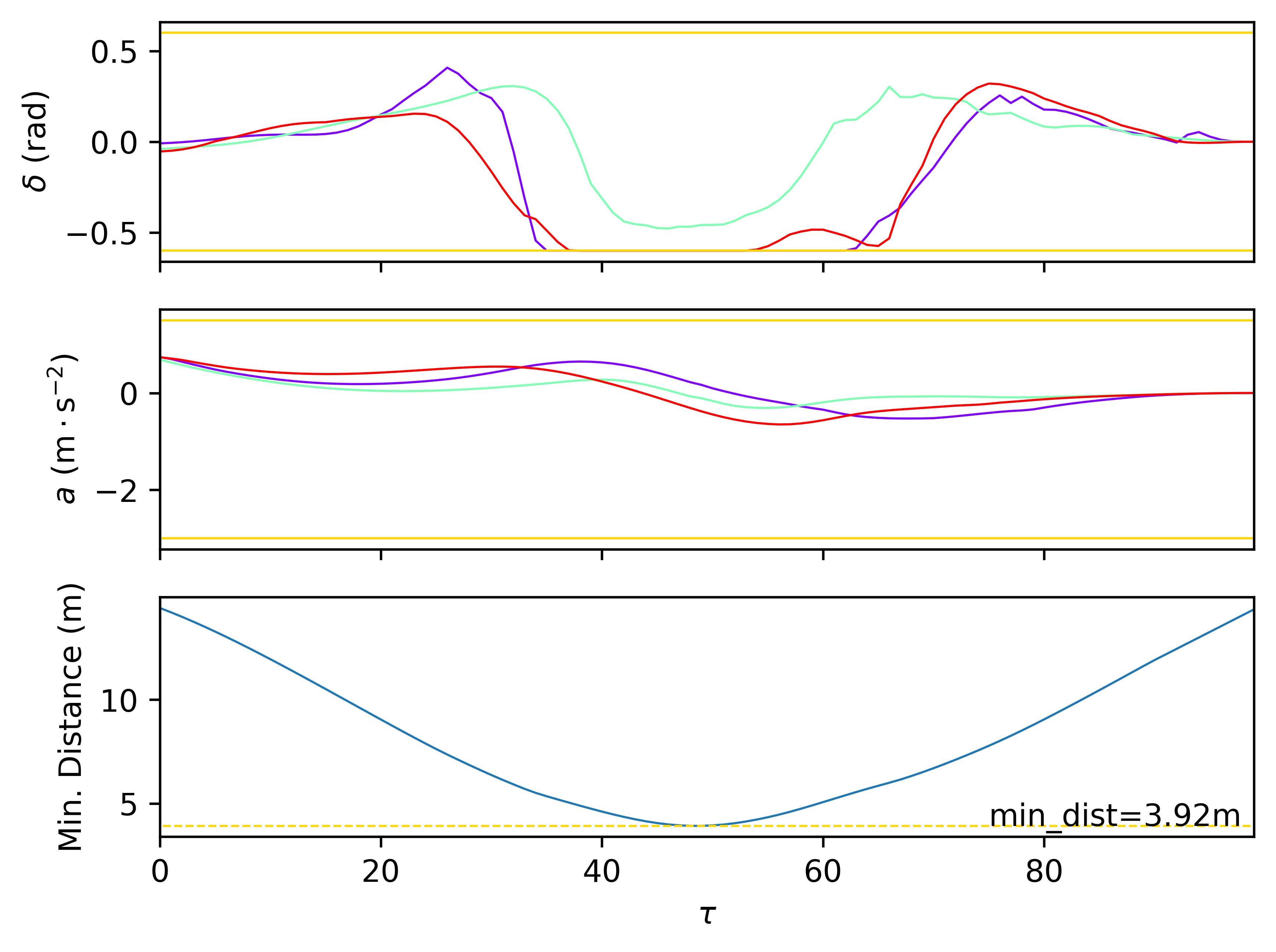}
\caption{Steering angles and accelerations of 3 vehicles and minimal distance between the center of vehicles in Scenario 1.}
\label{fig:scenario1Inputs}
\end{figure}

\begin{figure}[t]
\centering
\includegraphics[scale=0.50]{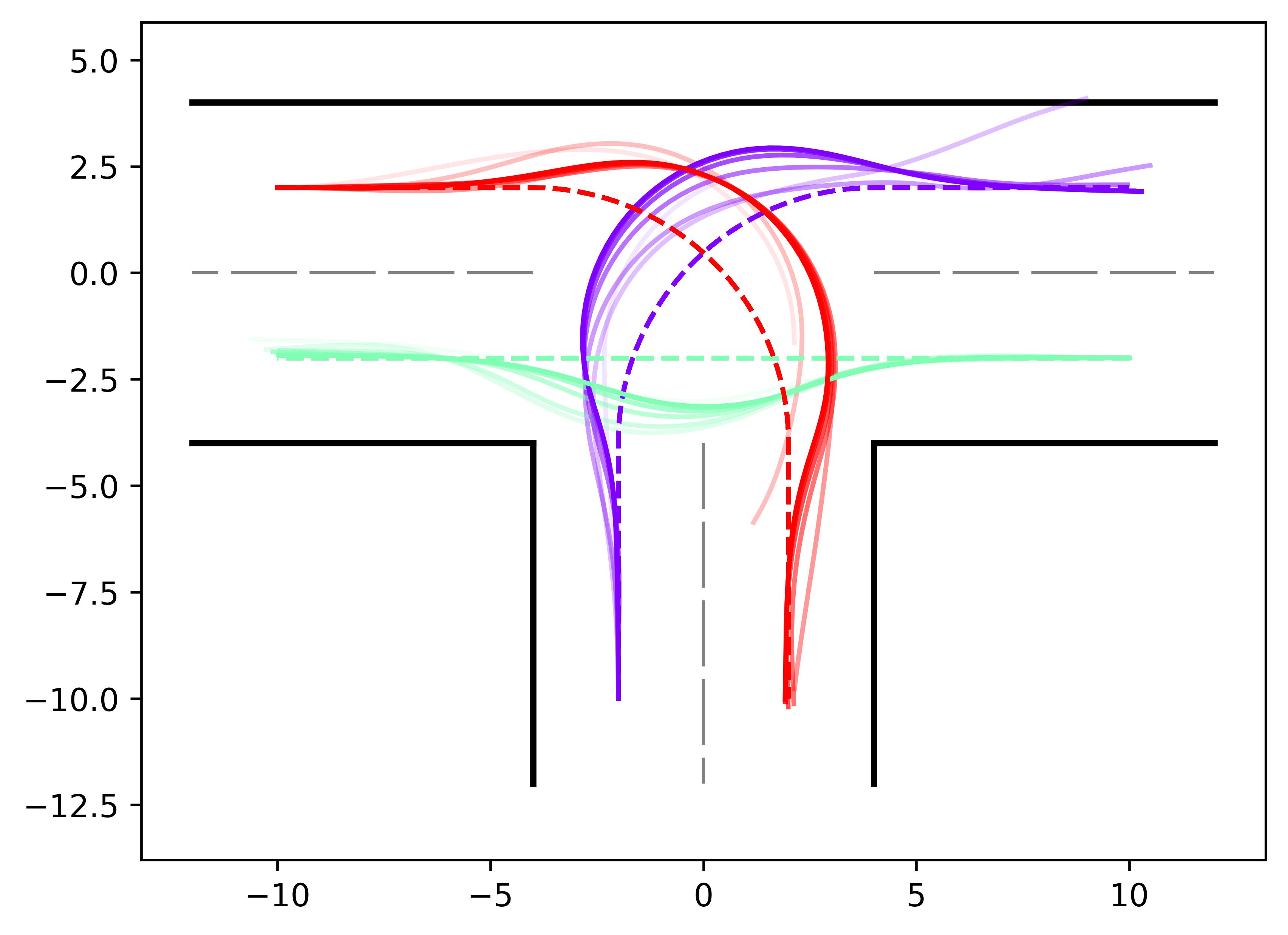}
\caption{Iterations of trajectories of 3 vehicles in Scenario 1.}
\label{fig:scenario1Iter}
\end{figure}

We first consider a T-junction scenario with three vehicles. As is shown in Fig. \ref{fig:scenario1Results}(a), the vehicles are represented by rectangles in different colors, and each dotted line is the reference trajectory corresponding to the vehicle of the same color, with each dot representing the reference position of the vehicle center corresponding to a particular time stamp.
The vehicles are required to pass the T-function following the reference trajectories as close as possible and avoid collisions.

For Scenario 1, we set $\sigma=0.1$ and $\rho = 0.01$. Due to the initialization scheme described in Remark \ref{initialization}, the stopping criterion of ADMM (Step 16 of Algorithm 2) can be set as termination after a fixed, small number of iterations. Here, we set the number of ADMM iterations to be 2. The results are shown in Fig. \ref{fig:scenario1Results}(b)-(d). Each sub-figure corresponds to a particular time stamp, with the rectangles showing the poses of the vehicles and the dotted lines showing their past trajectories. All three vehicles succeed in reaching their destination following smooth, feasible trajectories while keeping safe distances from each other to avoid collisions. Fig. \ref{fig:scenario1Inputs} shows the inputs and the minimal distance. Box constraints are satisfied and the minimal distance between the center of vehicles is $3.92\,\textup{m}$, which proves that the trajectories are collision-free. Both the single-process and the multi-process implementations yield numerically the same results.

\begin{figure}[t]
\centering
\includegraphics[scale=0.45]{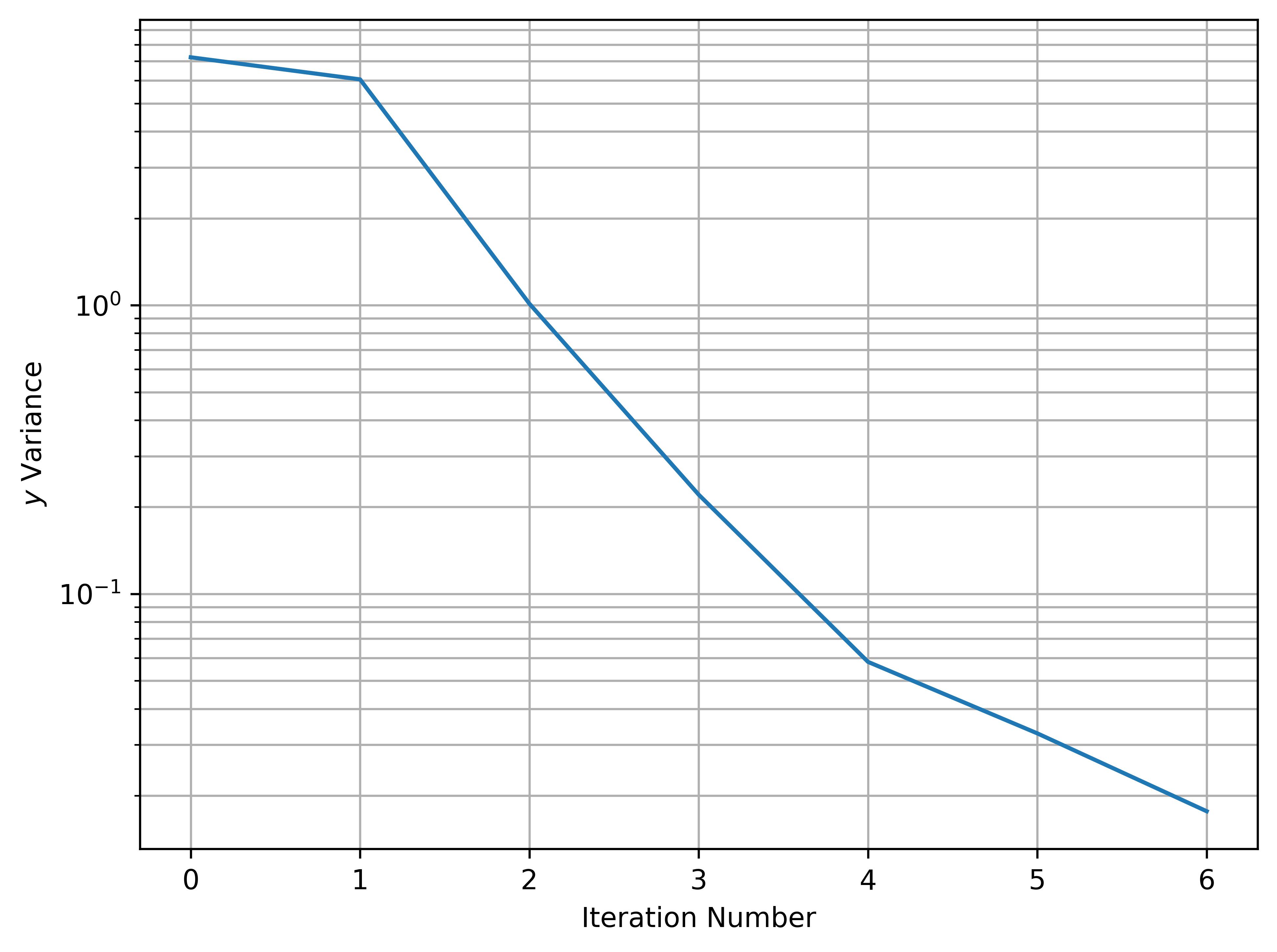}
\caption{Variance of $\{y^i\}$ with respect to number of iterations in Scenario 1.}
\label{fig:scenario1YVar}
\end{figure}

Our algorithm takes 7 iterations to stop. The drifting of vehicle trajectories through iterations are shown in Fig. \ref{fig:scenario1Iter}. The group of trajectories of the same color corresponds to the same vehicle, with more solid trajectories corresponding to a higher number of iterations. Fig. \ref{fig:scenario1Iter} qualitatively shows how the trajectories converge to the optimal trajectories through iterations. To further show the convergence of our algorithm, we adopt the conclusion that the set of dual variables $\{y^i\}$ should converge to the same vector $y^*$ as the algorithm converges. Therefore, we compute the variance of each element of $\{y^i\}$ over $i$ and obtain the mean of those variances. We plot the mean of variances with respect to the iteration number. The result is given in Fig. \ref{fig:scenario1YVar} in log-scale, showing monotonically decreasing of the variances of $\{y^i\}$, which clearly demonstrates the convergence of our algorithm.

The first row of Table \ref{Tab:time} demonstrates the computation time of all five implementations. The multi-process implementation of our algorithm takes $0.035\,\textup{s}$ to finish, which is the fastest of all. It reaches approximately $1.5\times$ speed compared to the centralized iLQR solver, $2.4\times$ speed to the single-process version of our method, $9.3\times$ speed to the IPOPT solver, and $23.4\times$ speed to the SQP solver. It should be noted that the single-process implementation of our proposed method is slower than the centralized iLQR solver, which implies that the proposed algorithm does not reduce the overall amount of computation, but provides a decentralized optimization framework such that parallel computing can be used to speed up the optimization process.

\begin{figure*}[t]
\centering
\subfigure[$\tau=0$]{\includegraphics[scale=0.41]{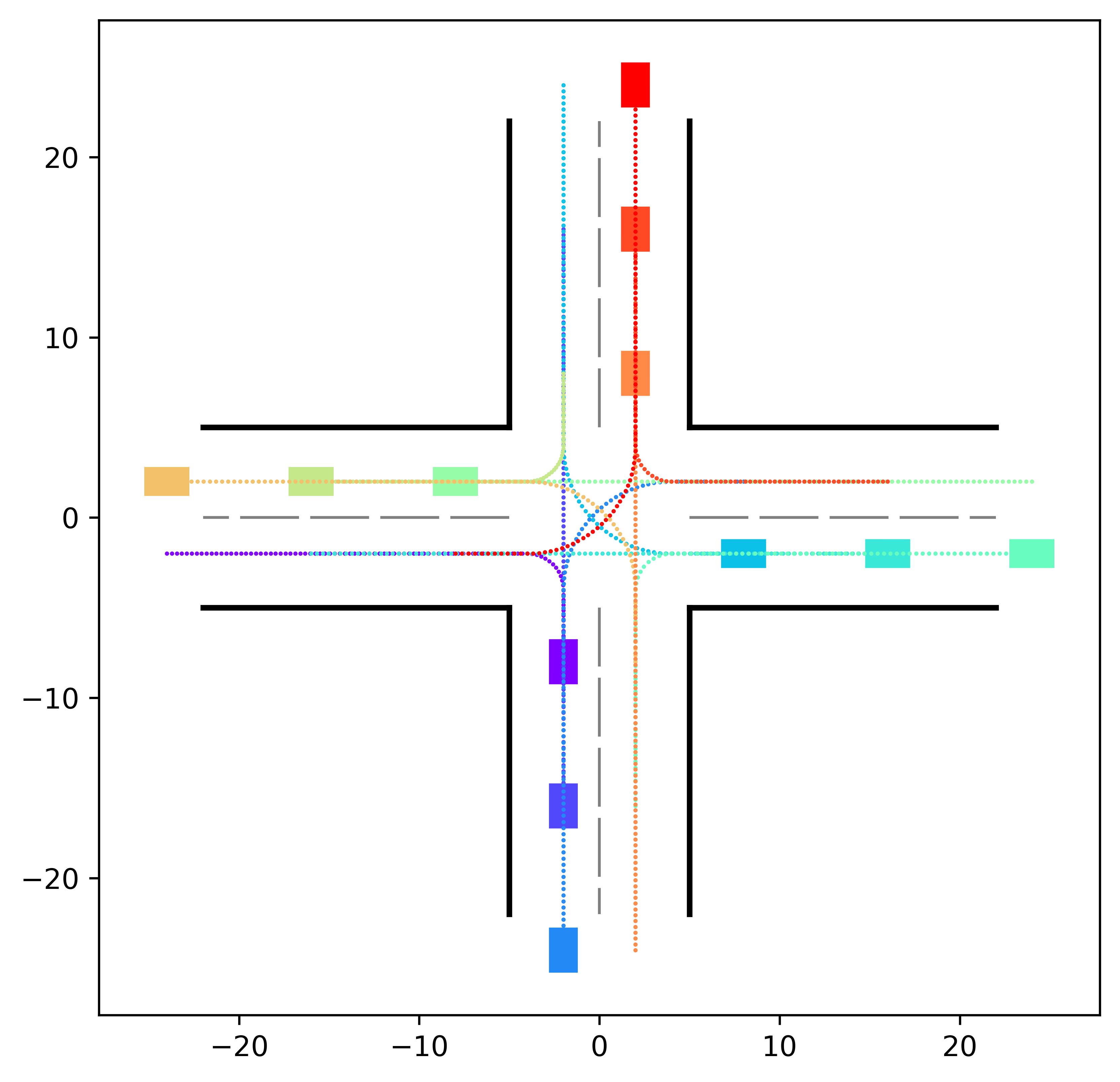}}
\subfigure[$\tau=25$]{\includegraphics[scale=0.41]{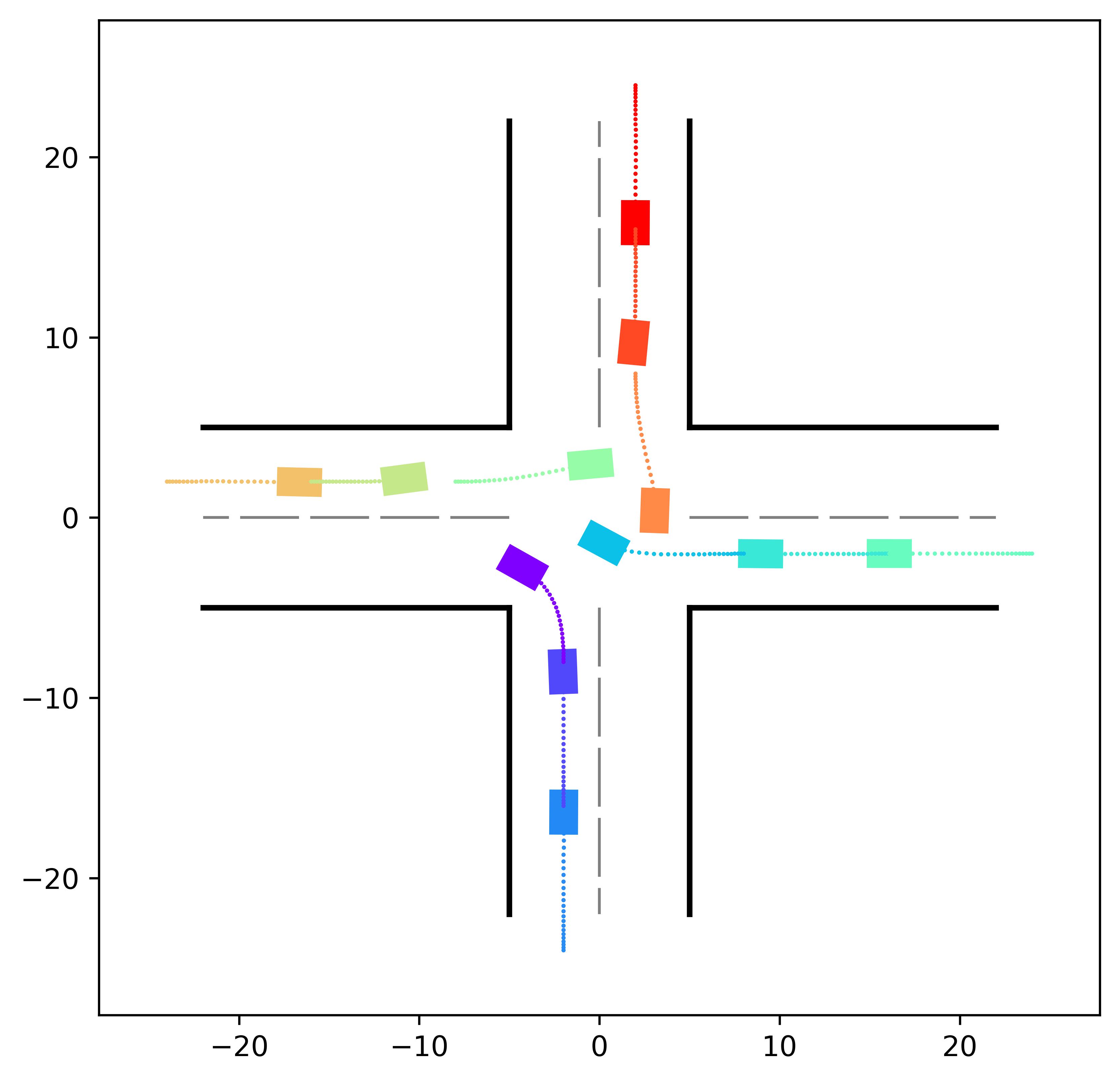}}
\subfigure[$\tau=50$]{\includegraphics[scale=0.41]{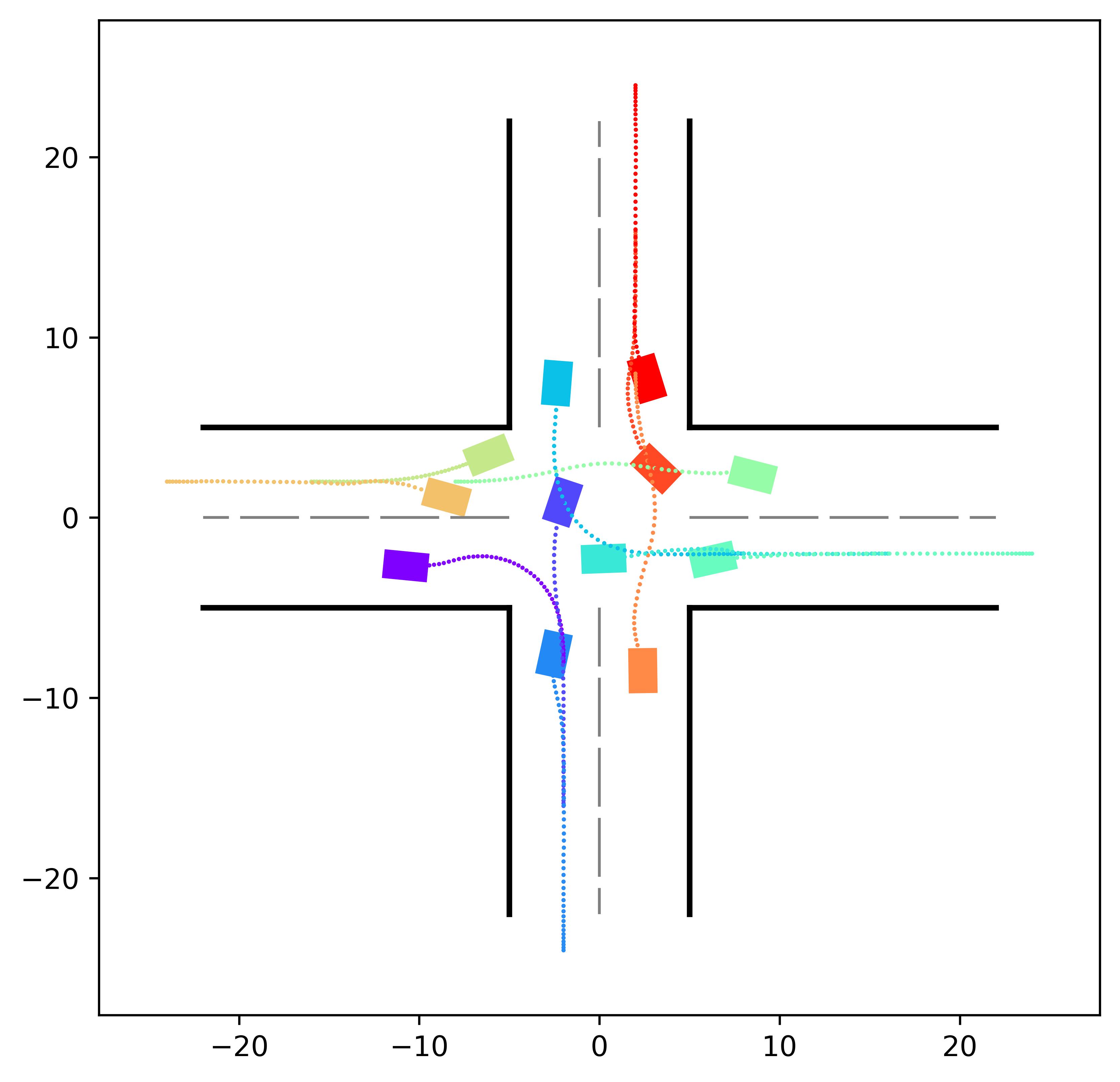}}

\centering
\subfigure[$\tau=60$]{\includegraphics[scale=0.41]{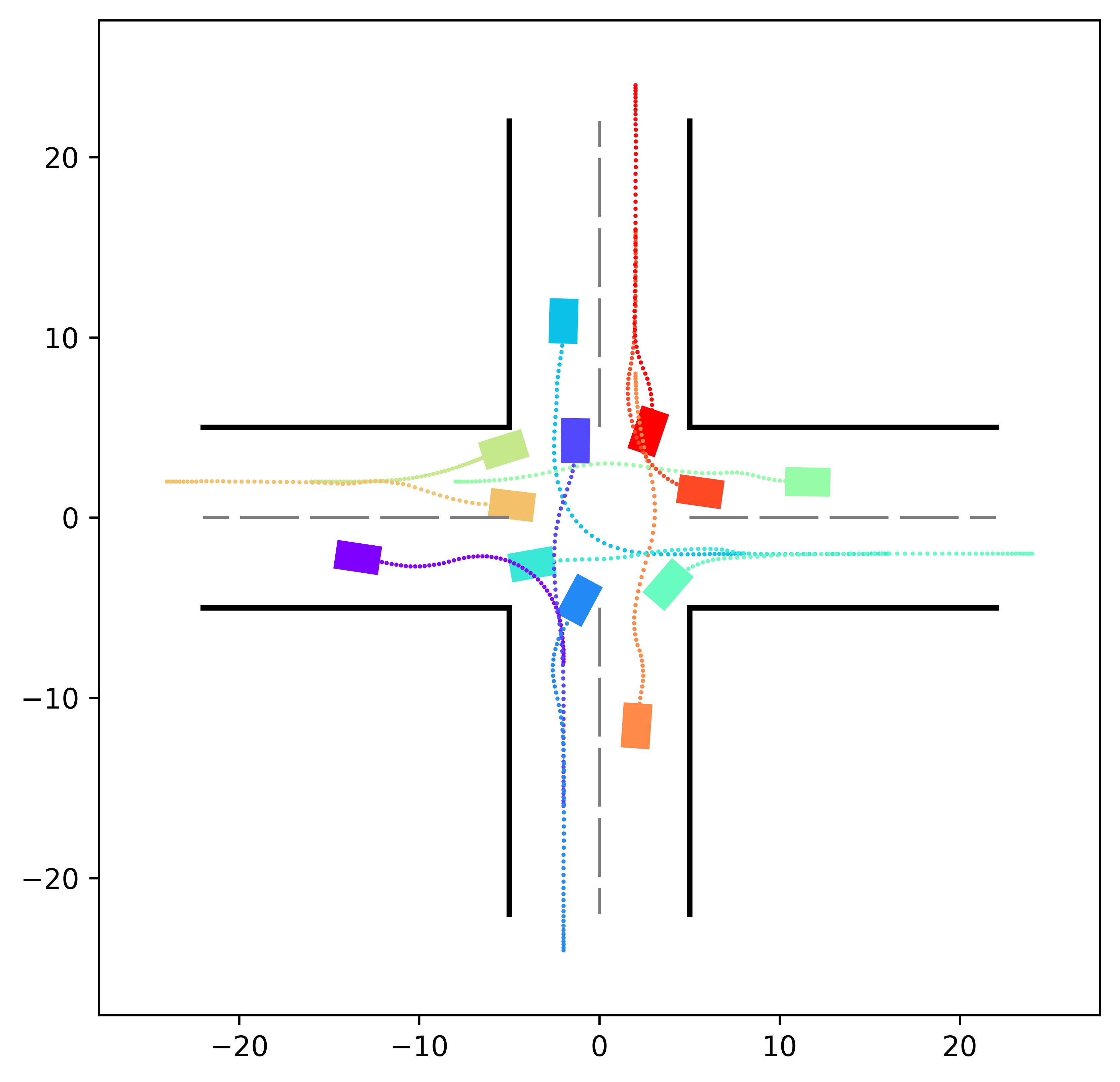}}
\subfigure[$\tau=74$]{\includegraphics[scale=0.41]{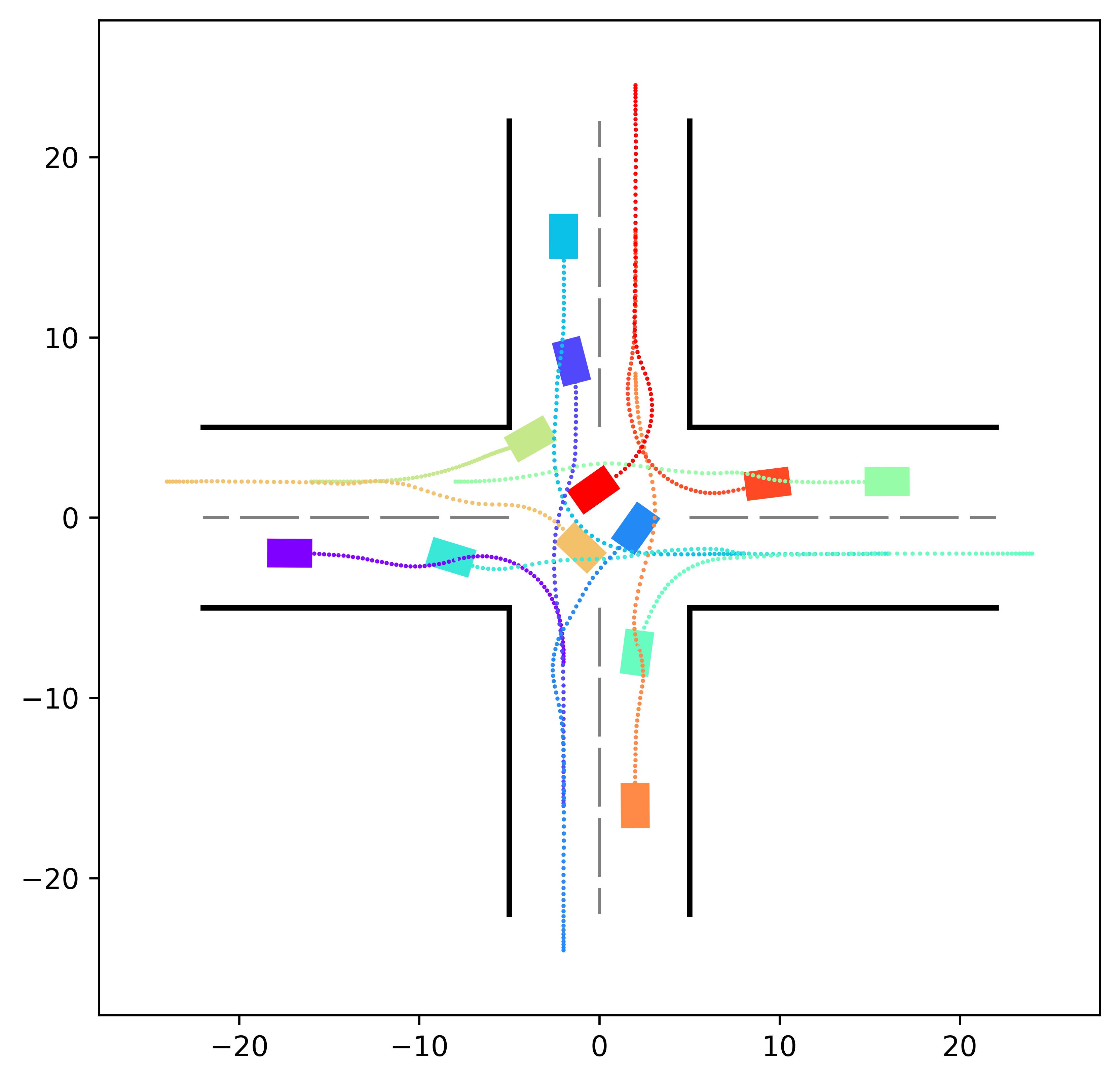}}
\subfigure[$\tau=99$]{\includegraphics[scale=0.41]{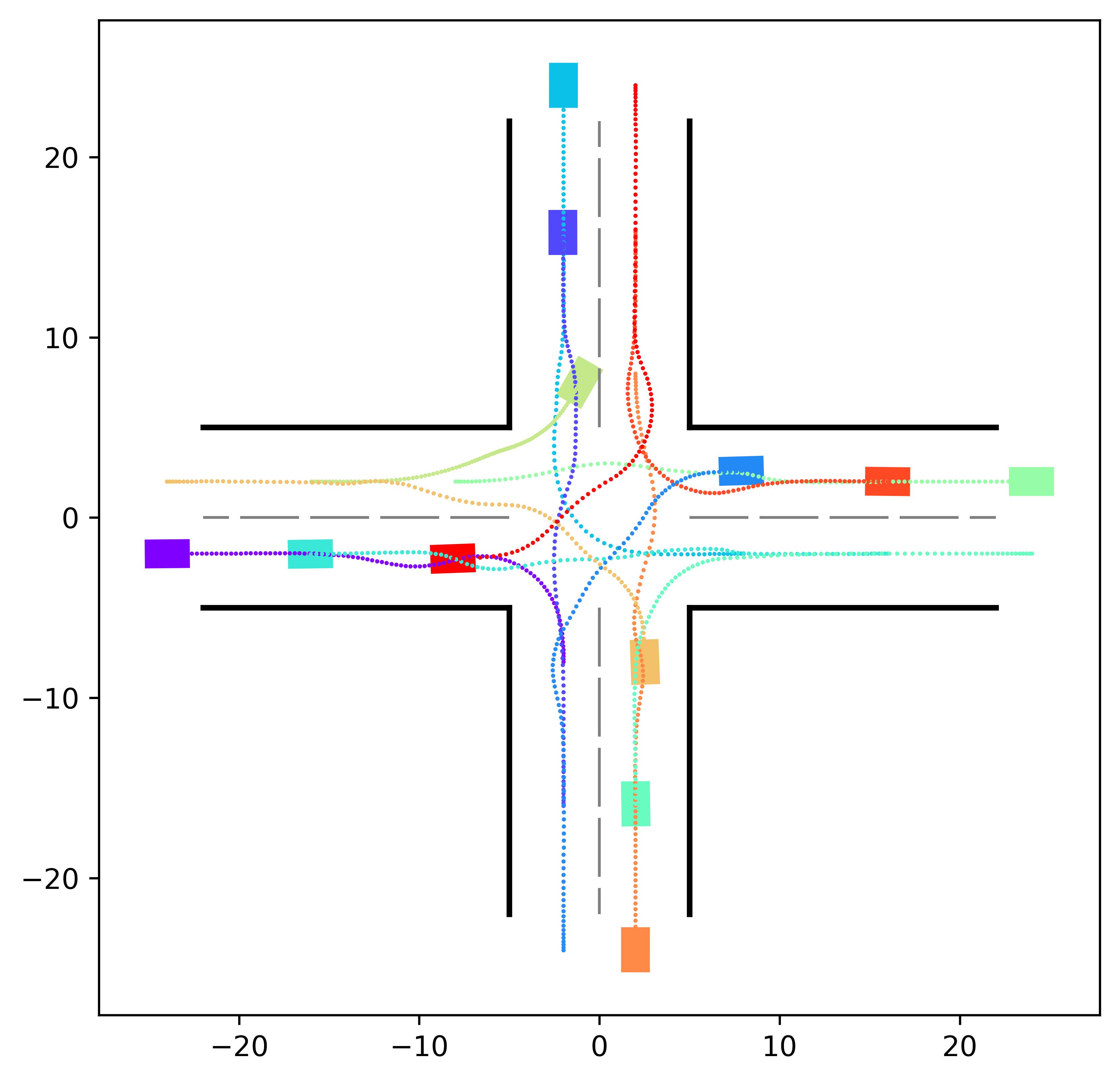}}

\caption{Simulation results for Scenario 2 with the proposed method at time stamps $\tau=0$, $\tau=25$, $\tau=50$, $\tau=60$, $\tau=74$, and $\tau=99$.}

\label{fig:scenario2Results}
\end{figure*}

\begin{figure}[h!]
\centering
\includegraphics[scale=0.50]{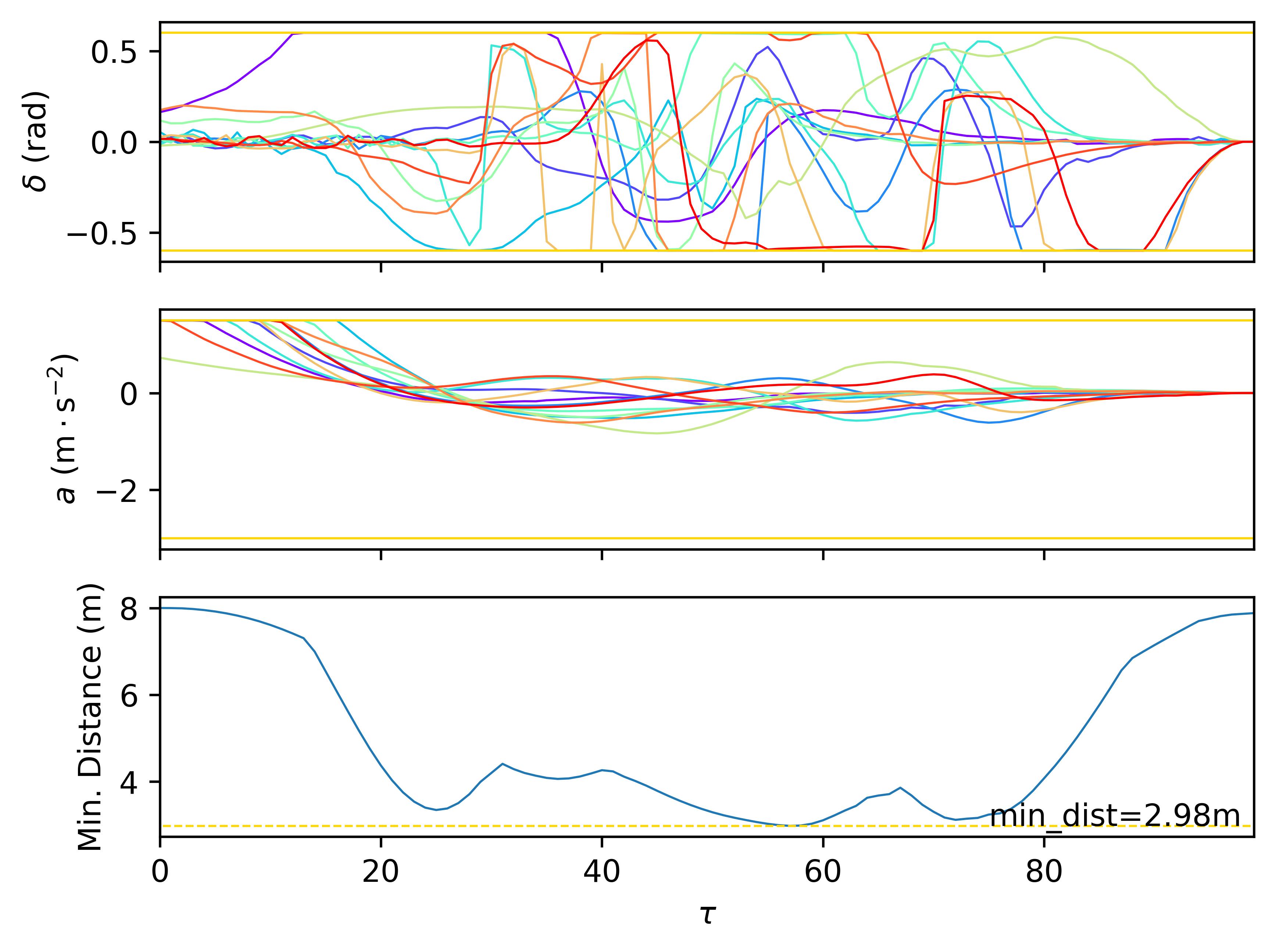}
\caption{Steering angles and accelerations of 12 vehicles and minimal distance between the center of vehicles in Scenario 2.}
\label{fig:scenario2Inputs}
\end{figure}

\begin{figure}[h]
\centering
\includegraphics[scale=0.4]{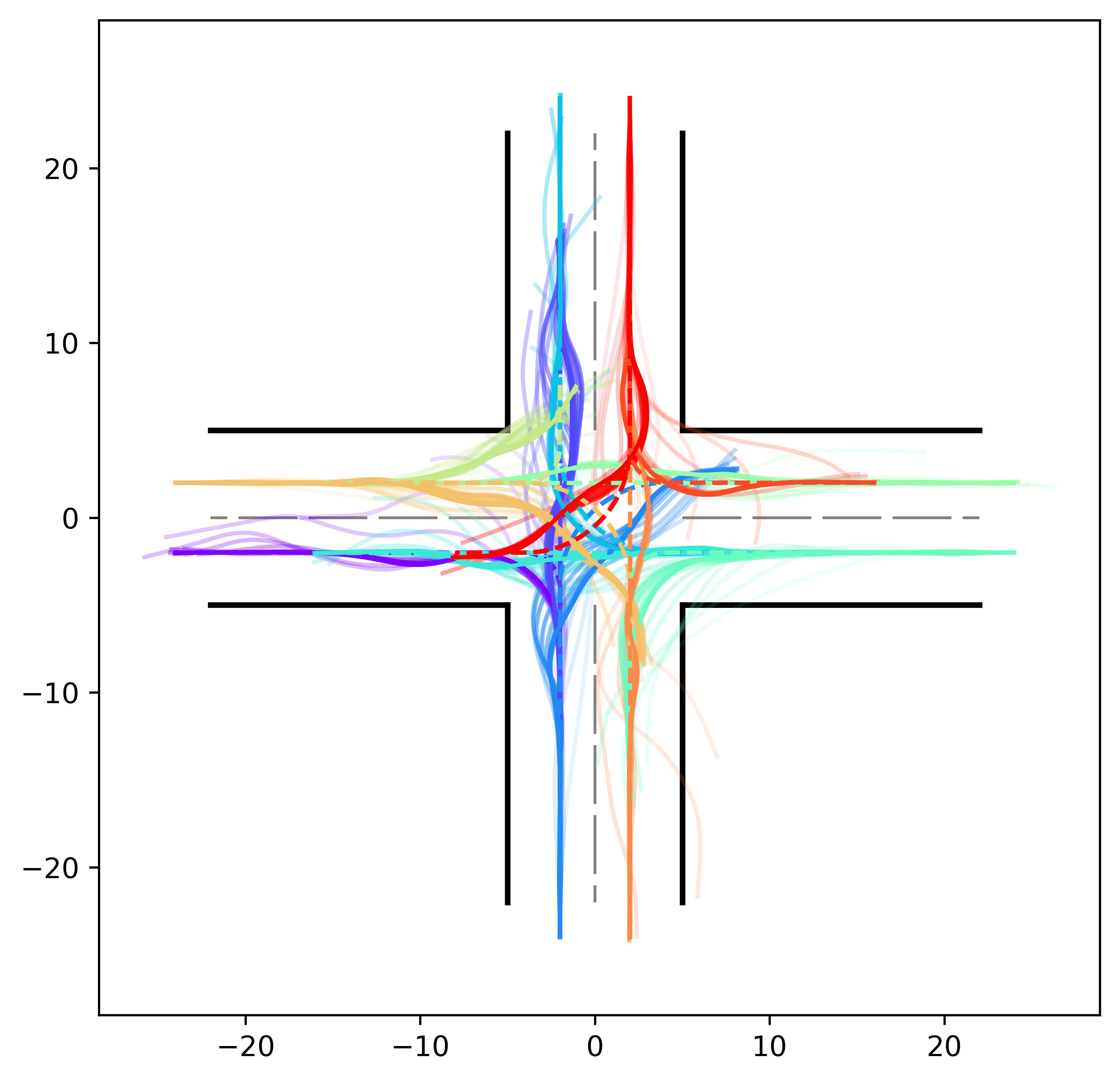}
\caption{Iterations of trajectories of 12 vehicles in Scenario 2.}
\label{fig:scenario2Iter}
\end{figure}

Due to different termination criteria, the quality of the final solution varies between solvers. Table \ref{Tab:quality} shows the overall cost of each solution. The overall cost of the proposed algorithm is nearly the same as the centralized iLQR solver and only slightly larger than the IPOPT solver, which implies comparable solution quality. Meanwhile, the SQP solver fails to obtain a collision-free solution, which is reflected by an overall cost that is much larger than the rest.

\begin{table*}[t]
\centering
\caption{Comparison of computation time between the proposed method, centralized iLQR, IPOPT, and SQP for Scenarios 1 and 2}
\begin{tabular}{lccccc}
\hline
\multirow{2}{*}{} & \multicolumn{2}{c}{Proposed method}                 & \multirow{2}{*}{Centralized iLQR} & \multirow{2}{*}{IPOPT} & \multirow{2}{*}{SQP} \\ \cline{2-3}
                  & Single-process & Multi-process                      &                                   &                        &                      \\ \hline
Scenario 1        & 0.084\,s       & \textbf{0.035\,s} & 0.052\,s                          & 0.327\,s               & 0.818\,s             \\
Scenario 2        & 1.691\,s       & \textbf{0.250\,s} & 1.186\,s                          & 11.140\,s              & ------               \\ \hline
\end{tabular}
\label{Tab:time}
\end{table*}

\subsubsection{Scenario 2}

\begin{figure}[t]
\centering
\includegraphics[scale=0.45]{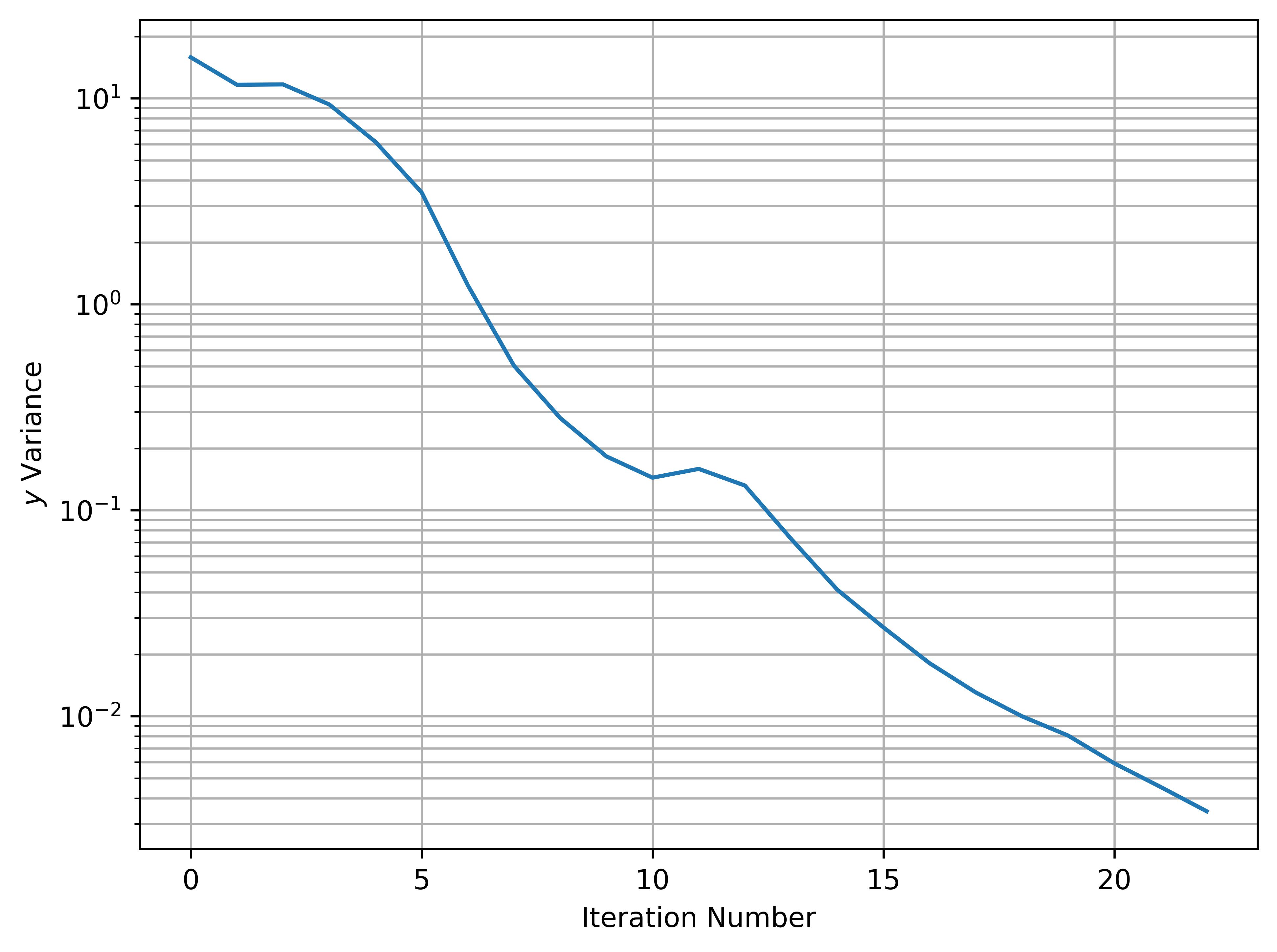}
\caption{Variance of $\{y^i\}$ with respect to number of iterations in Scenario 2.}
\label{fig:scenario2YVar}
\end{figure}

To further demonstrate the superiority in computational efficiency of our proposed method, we consider a scenario of a larger scale. As is shown in Fig. \ref{fig:scenario2Results}(a), Scenario 2 is an intersection with 12 vehicles passing simultaneously. We discover that for faster convergence, the parameters $\sigma$ and $\rho$ should be reduced with the number of vehicles increasing. Therefore, we scale down $\sigma$ and $\rho$ to $\sigma=0.01$ and $\rho=0.001$, and the number of ADMM iterations is set to be 3. The simulation results are shown in Fig. \ref{fig:scenario2Results}(b)-(f). Similarly, smooth trajectories are obtained. Fig. \ref{fig:scenario2Inputs} shows that the inputs satisfy box constraints and the minimal distance is $2.98\,\textup{m}$, which also guarantees safety. The proposed algorithm takes 24 iterations to finish. The iteration of trajectories is shown in Fig. \ref{fig:scenario2Iter}. Fig. \ref{fig:scenario2YVar} plots the variance of $\{y^i\}$ with respect to iteration number in log-scale. Again, the near monotonically decreasing of the variance of $\{y^i\}$ demonstrates the convergence.

The second row of Table \ref{Tab:time} shows the computation time of all implementations corresponding to Scenario 2. Again, the multi-process version of the proposed method is the fastest, taking $0.250s$ to finish, which is roughly $4.7\times$ as fast as the centralized iLQR solver, $6.8\times$ as fast as the single-process implementation of our method, and $44.6\times$ as fast as the IPOPT solver. For this scenario, the SQP solver fails to converge. Compared to scenario 1 with 3 vehicles, the proposed method obtains much higher folds of speed-up, which supports the claim that it scales better with the number of participating vehicles than baseline methods.

For the solution quality, the second row of Table \ref{Tab:quality} reveals that the solution of our method converges to a smaller overall cost than the centralized iLQR solver, and is comparable to the one obtained by the IPOPT solver, with the overall cost being slightly bigger. We consider that such a small compromise in the solution quality for a significant amount of speed-up is acceptable.

Although it is better to reduce the value of $\sigma$ and $\rho$ for the scenario of larger scale, keeping the same $\sigma$ and $\rho$ as Scenario 1 still yields satisfactory results with only slight suboptimality (see Table \ref{Tab:parameter}). The overall iteration number increases from 24 to 28, which causes the computation time to increase by roughly $20\%$.

\subsection{Discussion}

\begin{figure}[t]
\centering
\includegraphics[scale=0.5]{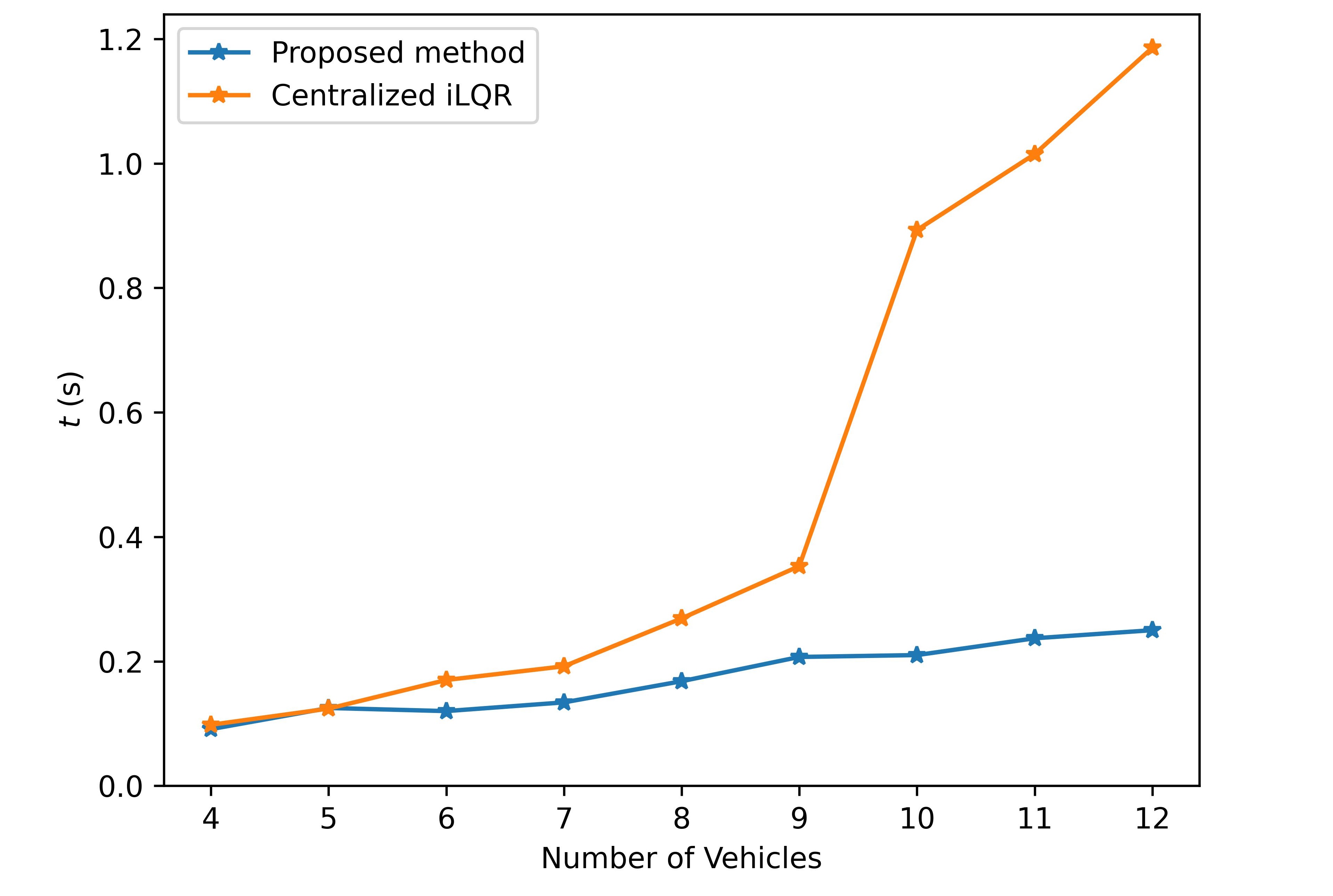}
\caption{Computation time with respect to different numbers of vehicles for the centralized iLQR solver and the proposed method.}
\label{fig:time}
\end{figure}

\begin{table}[h]
\centering
\caption{Comparison of performance for solving Scenario 2 with parameters $\sigma=0.01,\rho=0.001$ and $\sigma=0.1,\rho=0.01$}
\begin{tabular}{lccc}
\hline
                          & Iteration num.           & Cost          & Time    \\ \hline
$\sigma=0.01,\rho=0.001$  & 24             & 2303.7        & 0.250\,s \\
$\sigma=0.1,\rho=0.01$    & 28             & 2306.9        & 0.303\,s \\ \hline
                          &                &               &
\end{tabular}
\label{Tab:parameter}
\end{table}

\begin{table*}[t]
\centering
\caption{Comparison of final cost between the proposed method, centralized iLQR, IPOPT, and SQP for Scenarios 1 and 2}
\begin{tabular}{lccccc}
\hline
\multirow{2}{*}{} & \multicolumn{2}{c}{Proposed method}                 & \multirow{2}{*}{Centralized iLQR} & \multirow{2}{*}{IPOPT} & \multirow{2}{*}{SQP} \\ \cline{2-3}
                  & Single-process & Multi-process                      &                                   &                        &                      \\ \hline
Scenario 1 & 356.02 & 356.02         & 356.04        & \textbf{347.56}  & 626.40 \\
Scenario 2 & 2303.7 & 2303.7         & 2394.4        & \textbf{2297.7}  & ------ \\ \hline
           &        &                &               &                  &    
\end{tabular}
\label{Tab:quality}
\end{table*}

\subsubsection{Scalabilty}

To further show the scalability of the proposed algorithm, we conduct simulations on Scenario 2 with the number of vehicles varying from $N=4$ to $N=12$ while keeping all other parameters unchanged. For each case, we apply both the centralized iLQR solver and the proposed method with multi-process implementation and measure the computation time respectively. The results are shown in Fig. \ref{fig:time}. It is clear that when the number of vehicles is set to 4, our method takes nearly the same time as the centralized iLQR solver. However, as the number of vehicles increases, the computation time of the centralized iLQR method increases sharply and becomes much higher than our method. Quantitatively, when the number of CAVs increases by 3 times from $N=4$ to $N=12$, the computation time of centralized iLQR increases by $12.1\times$, which shows poor scalability. On the contrary, the computation time of our method increases by only $2.74\times$, which is slower than the increase of the number of vehicles.

\subsubsection{Safety}

In the proposed method, collision avoidance is achieved by adding a soft penalty to the overall cost whenever the distance between two vehicles is smaller than the given safe distance. Generally, such a formulation does not guarantee to produce a collision-free solution. However, safety conditions can still be satisfied by setting a large enough $\beta$ to scale up the collision avoidance penalty. Theoretically, when $\beta$ tends to infinity, the soft penalty essentially becomes hard constraint as an infinite cost is induced whenever two vehicles go within the safe distance, although an overly large $\beta$ could result in slow convergence.

\begin{figure}[t]
\centering
\subfigure[$\beta=1.00$ (collision)]{\includegraphics[scale=1.5]{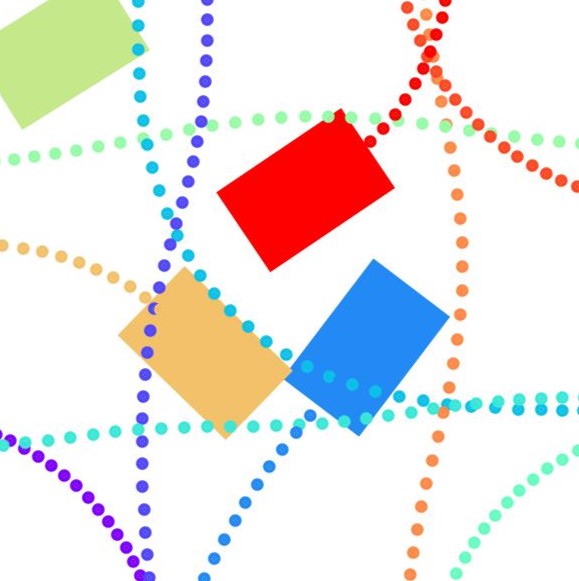}}
\subfigure[$\beta=1.44$ (safe)]{\includegraphics[scale=1.5]{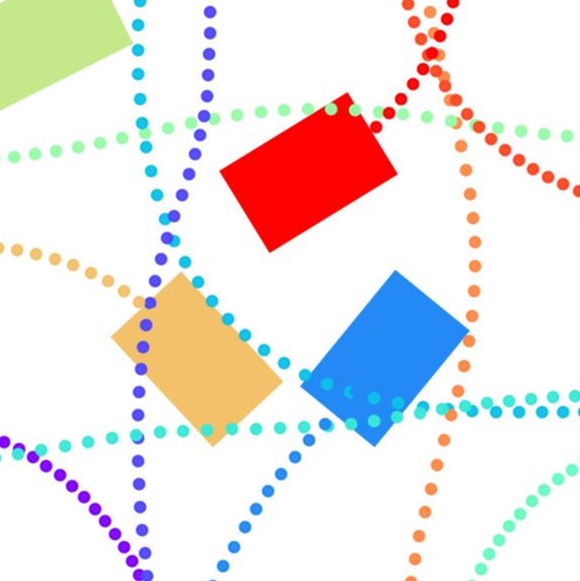}}

\centering
\subfigure[$\beta=1.96$ (safe)]{\includegraphics[scale=1.5]{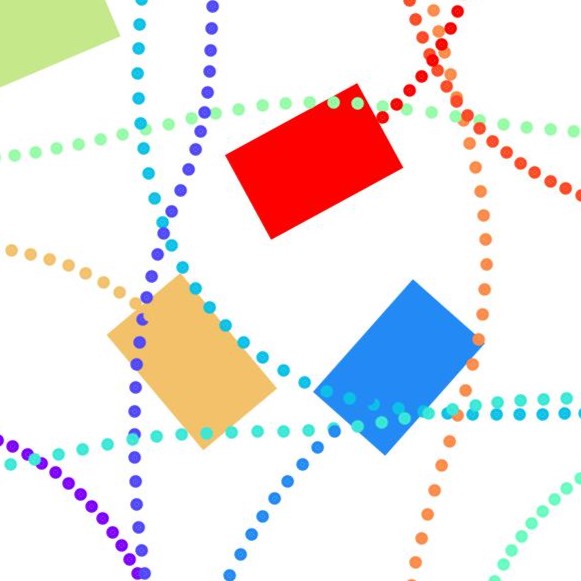}}
\subfigure[$\beta=2.56$ (safe)]{\includegraphics[scale=1.5]{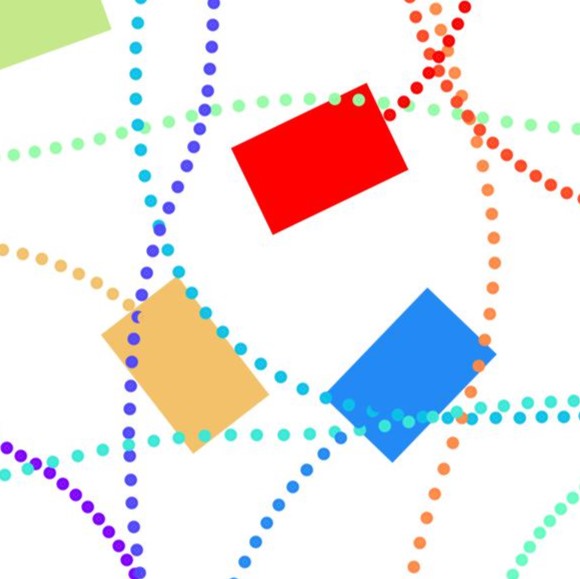}}
\centering
\caption{Simulation results for Scenario 2 with different values of $\beta$ at $\tau=72$.}
\label{fig:beta}
\end{figure}

To show this, we adapt different values of $\beta$ for solving Scenario 2 while keeping all other parameters unchanged, and we examine the effect of $\beta$ on both the minimal distance between vehicles and the overall computation time. The qualitative results at $\tau=72$ are shown in Fig. \ref{fig:beta}, which is the time when the closest distance between the corners of two vehicles occurs. It is clear that when $\beta=1.00$, the yellow car, and the blue car collide with each other. This collision is prevented when $\beta$ is set to be greater than $1.44$, and the spacing between the two vehicles continues to increase with increasing $\beta$. Quantitative results are presented in Table \ref{Tab:beta}. When $\beta$ increases from 1.00 to 2.56, the minimal distance between vehicles also increases, thus the safety is enhanced. However, a larger $\beta$ requires more iterations for the algorithm to converge, which results in a longer computation time.

Based on previous analysis, to ensure safety in real situations and maintain high computational efficiency, we can first initialize the algorithm with a reasonable value of $\beta$ and solve for a set of candidate trajectories. If collision is detected on the candidate trajectories, we increase $\beta$ to a higher value and solve for the trajectories again. We can keep looping for larger $\beta$ until collision-free trajectories are obtained.

\begin{table}[h]
\centering
\caption{Minimal distance between vehicle centers, iteration number, and computation time with varying $\beta$ for Scenario 2.}
\begin{tabular}{cccc}
\hline
$\beta$ & Minimal dis. & Iteration num. & Time   \\ \hline
1.00 & 2.59\,m        & 19       & 0.206\,s \\
1.21 & 2.81\,m        & 21       & 0.228\,s \\
1.44 & 2.98\,m        & 24       & 0.250\,s \\
1.69 & 3.15\,m        & 26       & 0.282\,s \\
1.96 & 3.30\,m        & 29       & 0.314\,s \\
2.25 & 3.37\,m        & 32       & 0.344\,s \\
2.56 & 3.43\,m        & 35       & 0.374\,s \\ \hline
\end{tabular}
\label{Tab:beta}
\end{table}

\section{Conclusion}

This work investigates the cooperative trajectory planning problem concerning multiple CAVs. The problem is formulated as a strongly non-convex optimization problem with non-linear vehicle dynamics and other pertinent constraints. We propose a decentralized optimization framework based on the dual consensus ADMM algorithm to distribute the computation load evenly among all CAVs, such that each CAV is iteratively solving an LQR problem with a fixed scale. We provide fully parallel implementation to enhance the efficiency and achieve real-time performance. Simulations on two traffic scenarios with the proposed method, centralized iLQR solver, IPOPT, and SQP are performed to validate the effectiveness and computational efficiency of the proposed method. Meanwhile, simulations on an increasing number of CAVs demonstrate the superiority of scalability of the proposed method compared to the centralized iLQR solver. A possible future work is to deploy our proposed algorithm on high-performance parallel processor (HPPP) such as GPU to scale up our algorithm to excessively large-scale scenarios containing hundreds of vehicles. Another future work is to perform field experiments to further substantiate the effectiveness of the proposed method.

\bibliographystyle{IEEEtran}
\bibliography{refs}

\end{document}